\documentclass[10pt,twocolumn,letterpaper]{article}

\usepackage{wacv}              

\usepackage{graphicx}
\usepackage{amsmath}
\usepackage{amssymb}
\usepackage{booktabs}

\usepackage{framed}
\usepackage{caption}
\usepackage{subcaption}
\usepackage{pgfplots}
\usepackage{esvect}

\usepackage{stackengine}
\usepackage{multirow}
\usepackage{tabularx}
\usepackage{wrapfig}
\usepackage{threeparttable}

\usepackage{tikz, xcolor}
\usetikzlibrary{calc,spy}
\usepackage{makecell}
\usepackage{overpic}

\usepackage[pagebackref,breaklinks,colorlinks]{hyperref}

\usepackage[capitalize]{cleveref}
\crefname{section}{Sec.}{Secs.}
\Crefname{section}{Section}{Sections}
\Crefname{table}{Table}{Tables}
\crefname{table}{Tab.}{Tabs.}

\def\wacvPaperID{374} 
\def\confName{WACV}
\def\confYear{2024}

\tikzstyle{closeup} = [
  opacity=1.0,          
  height=1.2cm,         
  width=1.2cm,          
  connect spies, white  
]

\tikzstyle{largewindow} = [white, line width=0.20mm]
\tikzstyle{smallwindow} = [white, line width=0.10mm]

\tikzstyle{largewindow_w} = [white, line width=0.50mm]
\tikzstyle{smallwindow_w} = [white, line width=0.15mm]
\tikzstyle{largewindow_b} = [blue, line width=0.50mm]
\tikzstyle{smallwindow_b} = [blue, line width=0.15mm]
\tikzstyle{smallwindow_c} = [red, line width=0.15mm]
\tikzstyle{closeup_b} = [
  opacity=1.0,          
  height=1cm,         
  width=1cm,          
  connect spies, blue  
]

\tikzstyle{closeup_c} = [
  opacity=1.0,          
  height=1.25cm,         
  width=1.25cm,          
  connect spies, red  
]

\tikzstyle{closeup_w} = [
  opacity=1.0,          
  height=1cm,         
  width=1cm,          
  connect spies, white  
]

\begin{document}

\title{Implicit Neural Image Stitching \\ With Enhanced and Blended Feature Reconstruction}

\author{
Minsu Kim$^1$, \;\; Jaewon Lee$^1$, \;\; Byeonghun Lee$^2$, \;\; Sunghoon Im$^1$, \; and \; Kyong Hwan Jin$^2$\thanks{Corresponding author.} \\
$^1$DGIST, Republic of Korea \; $^2$Korea University, Republic of Korea \\
$^1${\tt\small \{axin.kim, ljw3136, sunghoonim\}@dgist.ac.kr} \;
$^2${\tt\small \;\{byeonghun\_lee, kyong\_jin\}@korea.ac.kr} \\
}
\maketitle

\begin{abstract}
Existing frameworks for image stitching often provide visually reasonable stitchings. However, they suffer from blurry artifacts and disparities in illumination, depth level, etc. Although the recent learning-based stitchings relax such disparities, the required methods impose sacrifice of image qualities failing to capture high-frequency details for stitched images. To address the problem, we propose a novel approach, implicit Neural Image Stitching (NIS) that extends arbitrary-scale super-resolution. Our method estimates Fourier coefficients of images for quality-enhancing warps. Then, the suggested model blends color mismatches and misalignment in the latent space and decodes the features into RGB values of stitched images. Our experiments show that our approach achieves improvement in resolving the low-definition imaging of the previous deep image stitching with favorable accelerated image-enhancing methods. Our source code is available at \url{https://github.com/minshu-kim/NIS}.
\end{abstract}

\section{Introduction}
\label{sec:intro}
Image stitching aims to generate a wider field-of-view panorama from multiple scenes with arbitrary views. They provide rich visual information for various fields that require panoramic images including autonomous driving, virtual reality, and medical imaging.

Depending on the existence of a fixed grid transformation, image stitching is categorized as view-fixed \cite{wang2020multi, lai2019video, li2019attentive} or view-free methods\cite{gao2011constructing, lin2011smoothly, zaragoza2013projective, li2017parallax, nie2021unsupervised}. Previous view-free stitchings \cite{gao2011constructing, lin2011smoothly, zaragoza2013projective, li2017parallax, nie2021unsupervised, nie2020view, nie2020learning, nie2023learning} align multiple views without the priors of inter-relationship between given scenes. In contrast, view-fixed approaches \cite{wang2020multi, lai2019video, li2019attentive} use pre-defined grid transformations to stitch different views. Among them, a trainable stitching method \cite{nie2021unsupervised} with a neural network shows good qualities at capturing color mismatches and blending misalignment in latent space. Because the method is fast and automatic, it demonstrates its practicality for real-time applications like virtual reality (VR) \cite{anderson2016jump, kim2019deep}, autonomous driving \cite{wang2020multi, lai2019video}. However, the existing methods struggle with low image qualities caused by large deformations and a lack of constraints for reasonable trade-offs between image quality and blending.

\begin{figure}[t]
    \centering
    \includegraphics[width=0.95\linewidth]{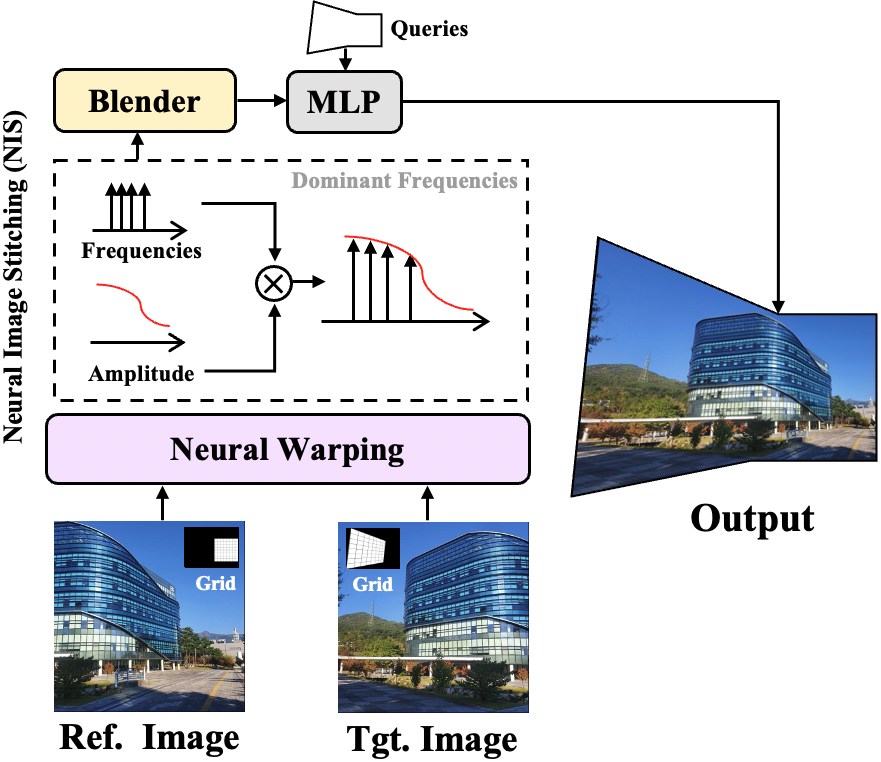}
    \vspace{-5pt}
    \caption{\textbf{Overview of our proposed image stitching.} Our model, NIS, estimates stitched images given a pair of target \& reference images and transformed grids. Neural warping extracts high-frequency-aware 2D features and aligns them according to the given grids. Blender takes the warped features and merges them into a feature map. A decoder (MLP) predicts an RGB signal at a coordinate $(x,y)$ of stitched image domain from a blended feature.}
    \label{fig:concept}
    \vspace{-15pt}
\end{figure}

\begin{figure*}[t]
\footnotesize
\centering
\stackunder[2pt]{
\begin{tikzpicture}[x=6cm, y=6cm, spy using outlines={every spy on node/.append style={smallwindow}}]
\node[anchor=south] (FigA) at (0,0) {\includegraphics[trim=0 100 0 0,clip,height=1.67in]{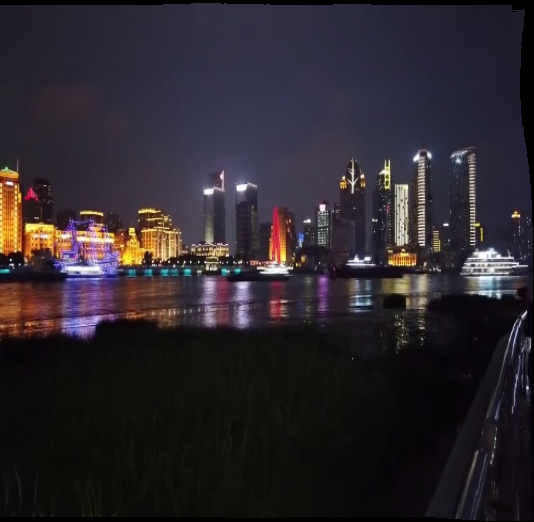}};
\spy [closeup,magnification=3] on ($(FigA)+(+0.334,+0.074)$) 
    in node[largewindow,anchor=east]       at ($(FigA.north) + (0.4,-0.13)$);
\spy [closeup,magnification=3] on ($(FigA)+( 0.148, 0.061)$)
    in node[largewindow,anchor=east]       at ($(FigA.north) + (0.12,-0.13)$);
\spy [closeup,magnification=1.7] on ($(FigA)+(-0.31,-0.073)$) 
    in node[largewindow,anchor=east]       at ($(FigA.north) + (-0.2,-0.13)$);
\end{tikzpicture}
}{APAP 
\cite{zaragoza2013projective}}
\hspace{-3mm}
\stackunder[2pt]{
\begin{tikzpicture}[x=6cm, y=6cm, spy using outlines={every spy on node/.append style={smallwindow}}]
\node[anchor=south] (FigA) at (0,0) {\includegraphics[trim=0 100 0 0,clip,height=1.67in]{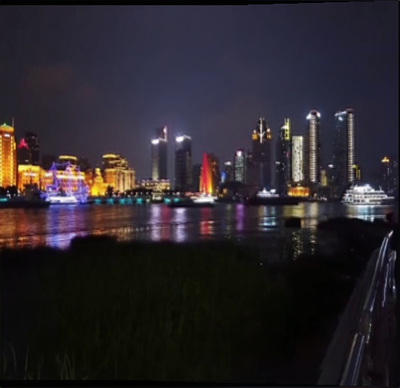}};
\spy [closeup,magnification=3] on ($(FigA)+(0.36, 0.056)$) 
    in node[largewindow,anchor=east]       at ($(FigA.north) + (0.4,-0.13)$);
\spy [closeup,magnification=3] on ($(FigA)+(0.147, 0.035)$)
    in node[largewindow,anchor=east]       at ($(FigA.north) + (0.12,-0.13)$);
\spy [closeup,magnification=1.7] on ($(FigA)+(-0.33,-0.11)$) 
    in node[largewindow,anchor=east]       at ($(FigA.north) + (-0.2,-0.13)$);
\end{tikzpicture}}
{UDIS \cite{nie2021unsupervised}}
\hspace{-2.4mm}
\stackunder[2pt]{
\begin{tikzpicture}[x=6cm, y=6cm, spy using outlines={every spy on node/.append style={smallwindow}}]
\node[anchor=south] (FigA) at (0,0) {\includegraphics[trim=0 100 0 0,clip,height=1.67in]{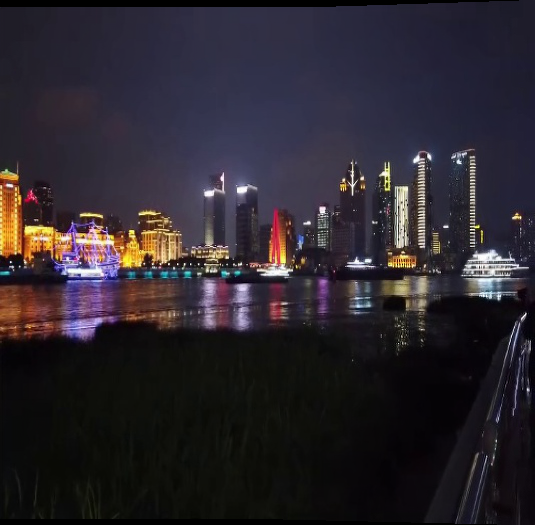}};
\spy [closeup,magnification=3] on ($(FigA)+(+0.33,+0.075)$) 
    in node[largewindow,anchor=east]       at ($(FigA.north) + (0.4,-0.13)$);
\spy [closeup,magnification=3.3] on ($(FigA)+( 0.145, 0.06)$)
    in node[largewindow,anchor=east]       at ($(FigA.north) + (0.12,-0.13)$);
\spy [closeup,magnification=1.7] on ($(FigA)+(-0.305,-0.072)$) 
    in node[largewindow,anchor=east]       at ($(FigA.north) + (-0.2,-0.13)$);
\end{tikzpicture}
}{\textit{Ours}}
\vspace*{-6pt}
\caption{\textbf{Visual Demonstration}. two-view image stitching with single perspective Homography.}
\vspace*{-10pt}
\label{fig:visual_demo}
\end{figure*}

The recent successful demonstrations of arbitrary-scale super-resolution (SR) with implicit neural representation (INR) \cite{chen2021learning, lee2022local} shed light on restoring such damaged low-definition images. Because the warp of an image using a grid is equivalent to the arbitrary-scale up-and-down sampling, the warped images can be enhanced by an extension of the arbitrary-scale SR. Following the idea, we propose a novel approach, implicit neural image stitching (NIS), which enables enhanced image stitching. With the assumption that aligning transformations are given, NIS predicts warped Fourier features with high-frequency details. Then, a CNN-based blender combines two aligned features into one latent variable in order to alleviate color mismatch and misalignment. Afterward, a decoder provides a representation of a stitched image over the extracted features. Our experiments show that the suggested INR restores local textures while keeping the estimation of a blended feature as the previous method \cite{nie2021unsupervised}. \\ 

In summary, our main contributions are as follows:
\begin{itemize}
\setlength{\leftmargin}{-.35in}
    \item We propose an implicit neural representation for image stitching that restores high-frequency details.

    \item We extend the concept of arbitrary-scale super-resolution into image stitching.
    \item Our model simplifies the image stitching pipeline performing various tasks into inference including warping paired images, blending misalignment and parallax errors, and relaxing blurred effects.
\end{itemize}

\section{Related Work}

\noindent{\bf Homography Estimation} 
The feature-based approaches estimate homography using the direct linear transformation (DLT) \cite{hartley2003multiple} given feature correspondences \cite{lowe2004distinctive, bay2006surf, rublee2011orb}.
Although those approaches can infer reasonable aligning transformations, they often fail to compute homography in challenging conditions where a pair of images is captured in different environments, such as day-to-night scenes and scenes with dynamic objects. To overcome the limitations, various deep homography estimators with superior feature extractors were suggested\cite{detone2016deep,nguyen2018unsupervised,erlik2017homography, le2020deep, zhou2019stn,zhang2020content,zhao2021deep,cao2022iterative, nie2020learning}. The first supervised and unsupervised deep homography estimators are proposed by Detone \textit{et al.} and Nguyen \textit{et al.} \cite{detone2016deep,nguyen2018unsupervised}, respectively. The proposed methods commonly estimate 4 corner displacements between two images. Then, using DLT algorithm and predicted displacements, they compute an aligning homography. Inspired by these works, various advanced methods are suggested, including variations of VGG-style networks \cite{erlik2017homography, le2020deep, zhou2019stn}, the moving content-aware model \cite{zhang2020content}, extracting one-channel of Lucas-Kanade feature map \cite{zhao2021deep}, and iterative architectures for inferencing a homography \cite{cao2022iterative, nie2020learning}.

\begin{figure*}[t]
\centering
\includegraphics[width=0.95\linewidth]{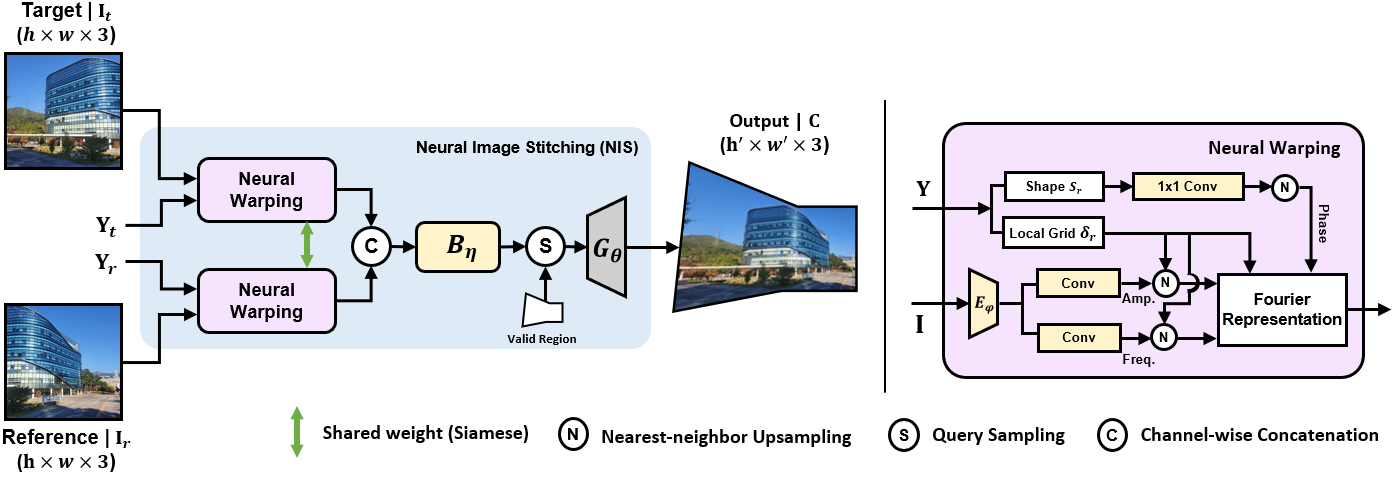}
\vspace{-8pt}
\caption{\textbf{Overall Flowchart of our Neural Image Stitching framework (NIS).} Our  NIS consists of a neural warping, a blender ($\boldsymbol{\it B_\eta}$), and a decoder ($\boldsymbol{\it G_\theta}$). We employ a homography estimator, IHN \cite{cao2022iterative}, for image alignment.}
\vspace{-10pt}
\label{fig:flowchart}
\end{figure*}

\noindent{\bf Implicit Neural Representation} Implicit neural representation \cite{hornik1989multilayer} approximates continuous signals such as 2D images and 3D shapes. Thanks to the property of INR, various tasks have been proposed including arbitrary image super-resolution (SR) \cite{chen2021learning, lee2022local}, SR for image warping \cite{son2021srwarp, yoon2021spheresr, ltew-jaewon-lee}, view synthesis \cite{mildenhall2021nerf}, etc. Among them, local INR \cite{chen2021learning, lee2022local, ltew-jaewon-lee} uses both feature maps from a CNN encoder and relative coordinates showing the robustness in generalization to out-of-scale datasets. Lee \textit{et al.} \cite{ltew-jaewon-lee} proposed an INR architecture, \textit{Local Texture Estimator for Warping} (LTEW). Compared to the previous work \cite{son2021srwarp}, it shows the robustness of the generalization performances by demonstrating the model under unseen grid transformations, including out-of-scale homographies, and \textit{Equirectangular projection} (ERP). Motivated by their successful application of arbitrary-scale SR for image warping, we propose an implicit neural function for image stitching.
\\
\noindent{\bf Image Blending} Blending techniques combines the overlapping regions of semantically aligned images as naturally as possible. There is a number of blending methods, including gradient-domain smoothing of color (a.k.a. Poisson blending) \cite{agarwala2004interactive, fattal2002gradient, perez2003poisson, uyttendaele2001eliminating}, alpha blending \cite{porter1984compositing}, multi-band blending \cite{zhang2014parallax}, and deep blending \cite{zhang2020deep, wu2019gp}. Among the approaches, the introduction of deep blending techniques paved the way for the previous learning-based image stitching \cite{nie2021unsupervised}.
\\
\noindent{\bf Image Stitching} There are two branches under image stitching, \textit{view-fixed} \cite{wang2020multi, lai2019video, li2019attentive, song2022weakly}, and \textit{view-free} tasks \cite{gao2011constructing, lin2011smoothly, zaragoza2013projective, li2017parallax, nie2021unsupervised, nie2020view, nie2020learning, jiang2022towards}. View-fixed scheme stitches images with given fixed views which are free from estimating an aligning transformation. 
View-fixed image stitching is often applied for specific tasks like autonomous driving \cite{wang2020multi, lai2019video} and surveillance videos \cite{li2019attentive} that use cameras with fixed locations. In contrast, view-free image stitching is applied to images with arbitrary views. It estimates geometric relations under dynamically distributed views. Then, it aligns the inputs and merges them. Gao \textit{et al.} \cite{gao2011constructing} proposed a model that estimates a dual homography to align two globally dominant frames in given images. Lin \textit{et al.} \cite{lin2011smoothly} proposed a method of estimating smoothly varying affine fields. Zaragoza \textit{et al.} \cite{zaragoza2013projective} suggested moving DLT to infer as-projective-as-possible (APAP) spatially weighted homography.
Li \textit{et al.} \cite{li2017parallax} computed \textit{thin plate splinte} (TPS) to finely optimize per-pixel deformations. Liao \textit{et al.} \cite{liao2019single} proposed single-perspective warps (SPW) and emphasizes image alignment using dual-feature (point + line) for structure-preserving image stitching. Jia \textit{et al.} \cite{jia2021leveraging} suggested leveraging line-point-consistence (LPC) which preserves the geometric structures of given scenes.
Recently, Nie \textit{et al.} \cite{nie2021unsupervised} proposed an unsupervised deep image stitching method (UDIS) which blends two warped images in latent space and decodes them to a stitched image. The approach shows favorable results in correcting illumination differences and relaxing parallax and misalignment. Inspired by such achievements, we propose a neural architecture for enhanced image stitching based on arbitrary-scale SR.

\section{Problem Formulation}
In our formulations, we define $\mathbf{x}$ and $\mathbf{y}\in \mathbb{R}^2$ as an input frame and a warped frame coordinate, respectively. $\mathbf{A}[\mathbf{x}]$ denotes the nearest neighbor interpolation for a signal $\mathbf{A}$ using a pixel coordinate. Given a reference $(\mathbf{I}_r)$ and a target image $(\mathbf{I}_t)$ where $\mathbf{I}_{\cdot}\in \mathbb{R}^{h\times w}$, we formulate an implicit neural function as
\begin{align}
    \mathcal{N}_{\Theta}: (\mathbf{y}_r,\mathbf{y}_t,\mathbf{I}_r[\mathbf{x}_r], \mathbf{I}_t[\mathbf{x}_t]) \mapsto (R,G,B).
\end{align}

The coordinates are obtained by transformation estimators \cite{cao2022iterative, li2017parallax} and NIS leverages them with input images. For detailed modularization of NIS, we decompose $\mathcal{N}_\Theta$ as
\begin{align}
    \mathcal{N}_\Theta = \mathbf{G}_\theta \circ \mathbf{B}_\eta \circ \boldsymbol{g},
\end{align}
where $\mathbf{G}_\theta, \mathbf{B}_\eta,$ and $\boldsymbol{g}$ are an implicit neural representation, a blender, and a displacement-dependent learnable warp, respectively. Our modular decomposition enables us to keep the property of the implicit neural representation that provides continuous RGB values of an implicitly stitched feature. Because the blending and reconstruction of high-frequency details are conflicting tasks, we suggest thorough strategies in \cref{sec:method}.

\subsection{Homography Estimation}\label{sec:homo_est}
As described in \cref{subsec: implementation}, we use a deep homography estimator IHN \cite{cao2022iterative} to train NIS. Because IHN recursively updates displacement vectors in the pre-fixed number of iterations, we design an additional formulation for it. Specifically, we formulate an unsupervised training for IHN \cite{cao2022iterative} as follows:

\vspace{-10pt}
\begin{gather}
    \small
    \hat{\mathbf{D}}=\arg\min_{\mathbf{D}}
    \sum_{k=1}^{K}\alpha^{K-k} \cdot \|\mathbf{I}_r-\mathit{W}(\mathbf{I}_t; \mathbf{H}_k)\|_1, \\
    \textrm{where} \quad \mathbf{H}_k=\mathit{f}_t(\mathbf{D}_k, \mathbf{c}), ~~ \mathbf{D}=\sum_{k}\mathbf{D}_k, \nonumber
    \label{eq:fourth}
\end{gather}
\noindent $\mathit{f}_t$ denotes the Direct Linear Transform (DLT)\cite{hartley2003multiple}. $\mathit{W}(A;B)$ means warping $A$ using a homography $B$. $K$ is the total number of iterations in IHN \cite{cao2022iterative}. $\mathbf{c}$ is a set of 4 corner coordinates of a target image. $\mathbf{D}_k$ and $\mathbf{H}_k$ are a k-th estimated displacement vector and a homography computed by Direct Linear Transform (DLT)\cite{hartley2003multiple}. $\alpha$ is the weight of the objective function set as 0.85 in our implementation. 

\begin{figure}[t]
    \centering
    \includegraphics[scale = 0.225]{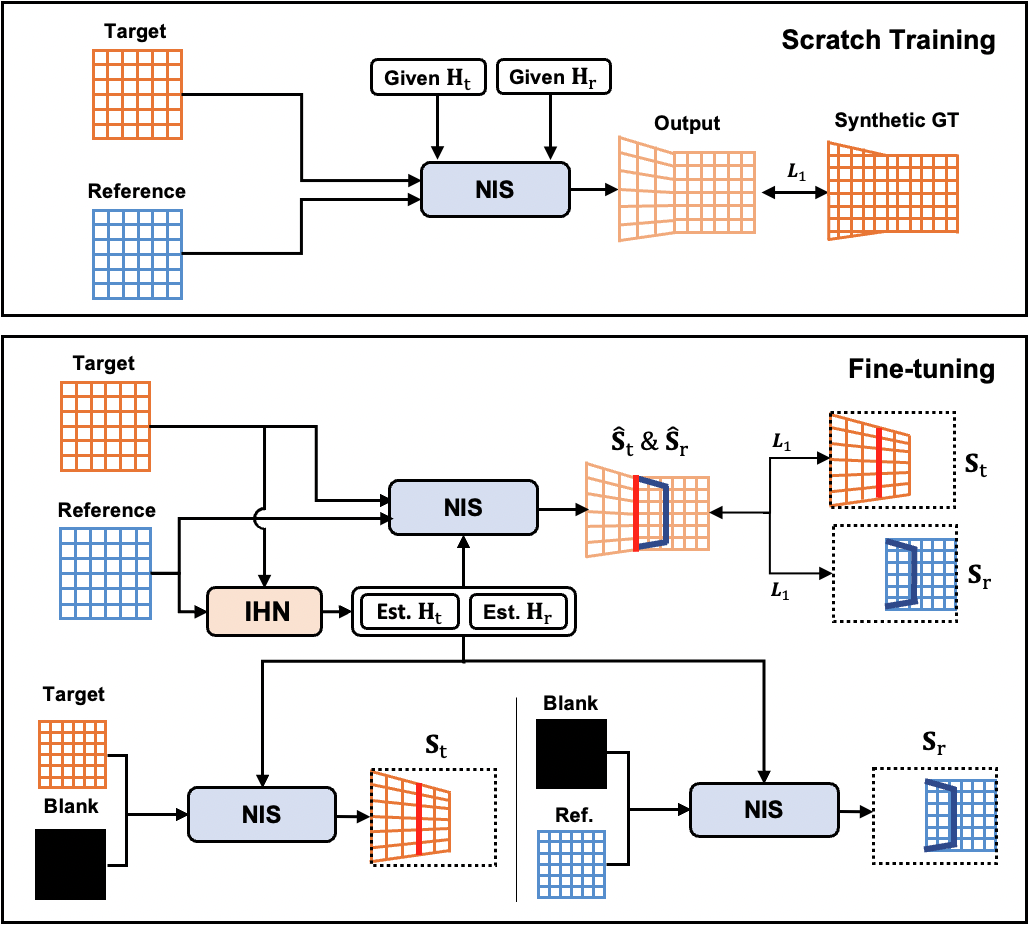}
    \caption{\textbf{Training Strategy.} In scratch training, we synthesize the aligning transformations and a stitched GT. In fine-tuning, we use IHN to estimate a homography $\mathbf{H}_t$ and train NIS in an unsupervised manner. Because there are no reference seam regions for $\hat{\mathbf S}_t$ and $\hat{\mathbf S}_r$, we estimate them with scratch-trained NIS.}
    \label{fig:strategy}
    \vspace{-10pt}
\end{figure}
\subsection{Implicit Neural Image Stitching} \label{sec:strategy}
\noindent\textbf{Estimation of Detail-aware Feature} We suggest \textit{Neural Warping} (NW, $\boldsymbol{g}$) for the estimation of warped features with enhanced textures. To this end, NW uses a vector containing direction and distance from the nearest referenceable coordinate. Specifically, we use a relative coordinate ($\mathbf{c}_m\in\mathbb{R}^2$) as a prior and a CNN-based filter that contains a learnable warping module restoring local textures in a warped feature $\mathbf{z} \in \mathbb{R}^{h'\times w'\times 4C}$ as
\begin{gather}
    \mathbf{z}[\mathbf{y}] = \boldsymbol{g}(\mathbf{E}_\varphi(\mathbf{I}^{IN})[\mathbf{x}], \mathbf{c}_m), \\
    \textrm{where} \;\; \mathbf{c}_m = \mathbf{y} - \mathbf{y}', \;\; \mathbf{y}' = \mathbf{X}[\mathbf{y}], \nonumber
\end{gather}
$\mathbf{E}_\varphi (\cdot): \mathbb{R}^3\mapsto \mathbb{R}^C$ is a CNN encoder \cite{lim2017enhanced}. $\mathbf{X}:=\{\mathbf{x}|\mathbf{x}\in\mathbb{R}^2\}$ is a grid of an input image $\mathbf{I}^{IN}\in\mathbb{R}^{h\times w}$ frame. 

\noindent\textbf{Fourier Coefficients for Stitching INR} Inspired by the successful demonstration of LTE \cite{lee2022local} for resolving spectral bias \cite{rahaman2019spectral}, NW leverages Fourier coefficients given a feature $\mathbf{z}'=\mathbf{E}_\varphi(\mathbf{I})\in\mathbb{R}^{h\times w\times C}$ and a cell $\mathbf{s}\in\mathbb{R}^{h'\times w'\times 10}$ that contains numerical Jacobian and Hessian tensors \cite{ltew-jaewon-lee} to predict the density distribution of a warped grid (or a cell decoding) as
\begin{align}
    \mathbf{A} = \boldsymbol{g}_a(\mathbf{z}'),\;\; \mathbf{F} = \boldsymbol{g}_f(\mathbf{z}'),\;\; \mathbf{P} = \boldsymbol{g}_p(\mathbf{s; \mathbf{y}}),
\end{align}
where $\boldsymbol{g}_a(\cdot)$ and $\boldsymbol{g}_f(\cdot):\mathbb{R}^C \mapsto \mathbb{R}^{4C}$ are an amplitude and a frequency estimator, respectively. $\boldsymbol{g}_p(\cdot): \mathbb{R}^{10} \mapsto \mathbb{R}^{2C}$ is a phase estimator. After the estimation of the coefficients, a detail-aware feature ($\mathbf{z}$) that represents a high-dimensional stitched image is associated as
\begin{gather}
    \mathbf{z}[\mathbf{y}] = \mathbf{A}[\mathbf{y}] \cdot (\cos \; \mathcal{F} + \sin \; \mathcal{F}), \\
    \textrm{where} \quad \mathcal{F} = \pi(\langle \mathbf{F}[\mathbf{\mathbf{y}}], \mathbf{c}_m \rangle + \mathbf{P}[\mathbf{y}]). \nonumber
\end{gather}

\noindent \textbf{Implicit Neural Representation} After NIS prepares a pair of features of a target ($\mathbf{z}_t$) and a reference ($\mathbf{z}_r$) images, it stitches them in a merged signal $\mathbf{C}'\in \mathbb{R}^{h'\times w'\in C}$. To this end, we design a blender $\mathbf{B}_\eta(\cdot, \cdot): (\mathbb{R}^{4C},\; \mathbb{R}^{4C}) \mapsto \mathbb{R}^C$ that learns a high-dimensional representation of a seamlessly stitched image as
\begin{align}
    \mathbf{C}'(\mathbf{y}) = \mathbf{B}_\eta(\mathbf{z}_t, \mathbf{z}_r).
\end{align}

Then a decoding INR $\mathbf{G_\theta}(\cdot): \mathbb{R}^{C} \mapsto \mathbb{R}^3$ takes a latent vector and provides an RGB of a stitched image ($\widehat{\mathbf{C}}$) as 
\begin{gather}
\widehat{\mathbf{C}}[\mathbf{y}_u] = \boldsymbol{\it G_\theta}(\mathbf{C}', \mathbf{y}_u),
\end{gather}
where $\mathbf{y}_u \in \mathbf{Y}_u \subset \mathbf{U}\;$ denotes a valid region coordinate, $\mathbf{U}:=[0,h)\times [0,w)$ is a uniform grid.

\vspace{8pt}
\section{Method} \label{sec:method}
NIS consists of a neural warping ($\boldsymbol{g}$), blender ($\mathbf{B}_\eta$), and a decoding INR ($\mathbf{G}_\theta$). The model is an end-to-end learnable model and infers a stitched image, automatically. This section provides detailed methods for training.
\subsection{Training Strategy}
We suggest 2 stage training to make the model strictly focus on each task (i.e., enhancing and blending). NIS learns high-frequency details in the first stage. After then, the model learns blended features in the second stage.  

\noindent\textbf{Learning Enhanced Details} Inspired by previous arbitrary-scale super-resolution \cite{chen2021learning, lee2022local, ltew-jaewon-lee, son2021srwarp}, we design supervised learning using data synthesize methods \cite{nie2020view}. We minimize L1 loss between ground truth and the estimated stitched images as
\begin{gather}
\mathcal L_{1}(\boldsymbol{\Theta})=\arg\min_{\boldsymbol{\Theta}} \; \sum_{\mathbf{y}_u} \|\mathbf{C}[\mathbf{y}_u]-\widehat{\mathbf{C}}[\mathbf{y}_u]\|_1. \label{eq:main}
\end{gather}

\noindent\textbf{Learning Blended Features}
The concatenation of two warped features causes inevitable differences between the distribution of overlapped and non-overlapped regions. The blending layer ($\mathbf{B}_\eta$) has to be trained to combine the two latent spaces into a single space correcting color mismatches and hiding parallax errors. To this end, the blender is constrained with a photometric seam loss \cite{nie2021unsupervised} as follows:
\begin{gather} \label{eq:seam}
        \mathcal L_{seam}(\boldsymbol{\Theta})=\arg\min_{\boldsymbol{\Theta}} \sum_n \|\hat{\mathbf S}_n - \mathbf{S}_n\|_1, \\
        \textrm{where} \;\; \hat{\mathbf S}_n=\mathbf{M}_n \odot \widehat{\mathbf{C}}, \nonumber \\
        \mathbf{S}_n=\mathbf{M}_n \odot \mathcal{N}_\Theta(\mathbf{y}_r, \mathbf{y}_t, \mathbf{I}_n, \vec{\mathbf{0}}), \;\; n\in \{r, t\}, \nonumber
\end{gather}
$\mathbf{M}_{(\cdot)}$ is a mask of seam regions between the target and reference. $\vec{\mathbf{0}} \in \mathbb{R}^{h\times w}$ is a blank image with the same size as an input image $\mathbf{I}_n$ domain. Overall procedures are provided in \cref{fig:strategy}.

\begin{table}[t]
    \centering
    \begin{subtable}[t]{\columnwidth}
        \centering
        \setlength{\tabcolsep}{1.5pt}
        \begin{tabular}{c
        |>{\centering\arraybackslash}p{0.26\textwidth}>{\centering\arraybackslash}p{0.26\textwidth}>{\centering\arraybackslash}p{0.25\textwidth}}
        \hline
        Method & mPSNR ($\uparrow$) & mSSIM ($\uparrow$) & \#Params. \\
        \hline\hline
        Bilinear & 34.78 & 0.96 & - \\
        Bicubic & \textcolor{blue}{36.25} & \textcolor{blue}{0.97} & -\\
        UDIS\cite{nie2021unsupervised} & 33.45 & \textcolor{blue}{0.97} & 8.0M \\
        NIS (\textit{ours})& \textcolor{red}{38.69} & \textcolor{red}{0.98} & 3.2M \\
        \hline
        \end{tabular}
        \subcaption{Evaluation on Synthetic Images.}
        \label{tab:Quan_stit_syn}
    \end{subtable}
    \\
    \vspace{5pt}
    \begin{subtable}[t]{\columnwidth}
        \centering
        \setlength{\tabcolsep}{1.5pt}
        \begin{tabular}{c
        |>{\centering\arraybackslash}p{0.22\textwidth}>{\centering\arraybackslash}p{0.22\textwidth}>{\centering\arraybackslash}p{0.22\textwidth}}
        \hline
        Method & NIQE ($\downarrow$) & PIQE ($\downarrow$) & BRISQUE ($\downarrow$)\\
        \hline\hline
        APAP\cite{zaragoza2013projective} & 3.30 & 46.95 & 34.72 \\
        Robust ELA\cite{li2017parallax} & 3.59 & 53.67 & 37.78 \\
        SPW\cite{liao2019single} & 3.39 & 51.68 & 36.84 \\
        LPC\cite{jia2021leveraging} & 3.37 & 50.81 & 37.15 \\
        LPC + Graph Cut & 3.50 & 50.63 & 37.14 \\
        UDIS\cite{nie2021unsupervised}& 3.43 & 50.01 & 36.71 \\
        IHN\cite{cao2022iterative}+NIS & \textcolor{blue}{3.28} & \textcolor{blue}{46.21} & \textcolor{blue}{33.17} \\
        IHN\cite{cao2022iterative}+NIS (F) & \textcolor{red}{3.15} & \textcolor{red}{43.05} & \textcolor{red}{31.14} \\
        \hline
        \end{tabular}
        \subcaption{Evaluation on Real Images.}
        \label{tab:Quan_stit_real}
        \vspace{-5pt}
    \end{subtable}
    \caption{\textbf{Quantitative Comparison} on Stitching performance. \textcolor{red}{Red} and \textcolor{blue}{Blue} colors indicate the best and the second-best performances, respectively.}
    \vspace{-10pt}
\end{table}

\subsection{Training Details}
\noindent{\bf Enhancement}
We apply an on-the-fly data generation that warps a mini-batch with the same homography. It provides synthesized paired images and ground truth stitched images per a mini-batch. To generate synthetic datasets, we follow the whole pipeline of \cite{nie2020view}. We set the ratio of 4 corner offsets $(\Delta_x, \Delta_y)$ as random values within 25\% of cropped image resolutions $h_c, w_c$. For training, we randomly sample M spatial query points from valid regions and minimize the $L_1$ loss between RGB prediction and ground truth as in \cite{ltew-jaewon-lee}. This augmentation strategy is as follows:
\begin{gather} \label{eq:otf}
        (\Delta_x, \Delta_y) = (a_x\cdot w_c,\; a_y\cdot h_c) \in \mathbb{R}^2, \\
        \textrm{where}\; a_l \sim \mathcal{U}(0, 0.25) \;\; x, y \in l. \nonumber
\end{gather}
\noindent{\bf Blending}
In this stage, NIS learns to correct color mismatches and misalignments. Because there is no given transformation prior in the real images, we use a homography estimator IHN \cite{cao2022iterative}, which was trained with the suggested unsupervised training methods  \cite{nie2021unsupervised}. Our network is trained only with the queries extracted from the seam regions of two images.
We freeze NIS except for the decoding INR and fine-tune it as a representation provider for blended stitched images. Because no ground truth or reference exists for predicted seam region RGBs, we generate warped target and reference images to use them as reference images. By forwarding an image (target or reference image) and a blank image into a frozen, scratch-trained NIS, we get the warped images and use them as references for predicted samples. The elements of a blank image are set to 0.

\subsection{Training Configurations}
In the first stage of training, we train NIS with a batch size of 20 and 250,000 iterations. We choose the crop size $h_c\times w_c$ as $48\times48$.
In the second stage, we fine-tune the model with a single batch size. We set the number of iterations as 60,000 and 300,000, respectively. For the second stage, the resolution of input images is downsampled to $128\times 128$ by bilinear downsampling. We use Adam optimizer \cite{kingma2014adam} with $\beta_1=0.9$ and $\beta_2=0.999$. The learning rates are initialized as $1\times10^{-4}$ for the first and second stages. The learning rates are exponentially decayed by a factor of 0.98 for every epoch.

\begin{table}[t]
    \centering
    \setlength{\tabcolsep}{1.3pt}
    \scriptsize
    \begin{tabular}{
        c>{\centering\arraybackslash}p{0.8cm}>{\centering\arraybackslash}p{1.8cm}>{\centering\arraybackslash}p{0.6cm}>{\centering\arraybackslash}p{1cm}>{\centering\arraybackslash}p{2cm}
    }
        \Xhline{2\arrayrulewidth}
        \multirow{2}{*}{Method} & \multicolumn{2}{c}{Size} & Mem. & Time & \multirow{2}{*}{Resource}\\
        \cline{2-3}
         & Input & Stitched & (GB) & (ms) & \\
        \hline
        \multirow{3}{*}{LPC\cite{jia2021leveraging}}
        & $128^2$ & $192^2$ & - & 109 & AMD Ryzen \\
        & $512^2$ & $784^2$ & - & 4,883 & 4800H\\
        & $1024^2$ & $1536^2$ & - & 28,446 & (CPU)\\
        \hline
        \multirow{3}{*}{{UDIS$^{\dagger}$}\cite{nie2021unsupervised}}
        & $128^2$ & $192^2$ & 9.1 & 74 & NVIDIA \\
        & $512^2$ & $784^2$ & 9.3 & 128 & RTX 3090\\
        & $1024^2$ & $1536^2$ & 9.7 & 253 & (GPU)\\
        \hline
        \multirow{3}{*}{NIS}
        & $128^2$ & $192^2$ & 0.5 & 58 & NVIDIA \\ 
        & $512^2$ & $784^2$ & 5.4 & 384 & RTX 3090  \\
        & $1024^2$ & $1536^2$ & 17.3 & 1,944 & (GPU)\\
        \Xhline{2\arrayrulewidth}
    \end{tabular}
    \vspace{-8pt}
    \caption{\textbf{Specifications of Stitching Methods.} $^\dagger$ denotes a third-party implementation.}
    \label{tab:time}
    \vspace{-10pt}   
\end{table}

\begin{figure*}[t]
\footnotesize
\centering
\begin{tikzpicture}[x=6cm, y=6cm, spy using outlines={every spy on node/.append style={smallwindow_w}}]
\node[anchor=south] (FigA) at (0,0) {\includegraphics[height=1.62in,width=1.62in]{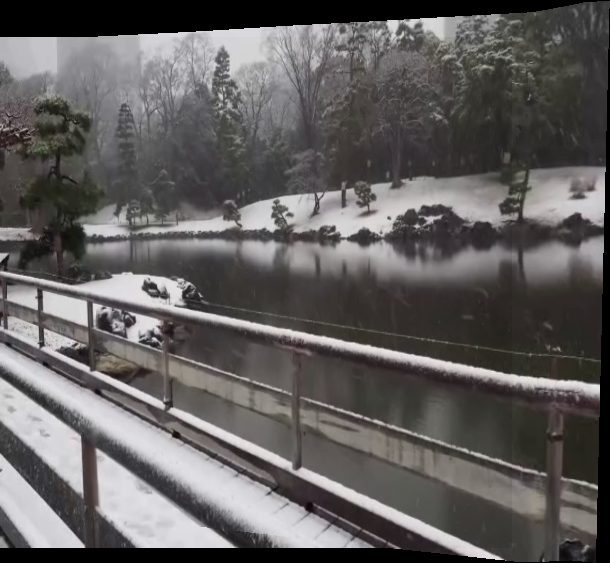}};
\spy [closeup_w,magnification=2.5] on ($(FigA)+(0.00, -0.070)$) 
    in node[largewindow_w,anchor=east]       at ($(FigA.north) + (0.33,-0.61)$);
\spy [closeup_w,magnification=3.5] on ($(FigA)+(-0.212, -0.045)$) 
    in node[largewindow_w,anchor=east]       at ($(FigA.north) + (-0.17,-0.61)$);
\end{tikzpicture}
\begin{tikzpicture}[x=6cm, y=6cm, spy using outlines={every spy on node/.append style={smallwindow_w}}]
\node[anchor=south] (FigA) at (0,0) {\includegraphics[height=1.62in,width=1.62in]{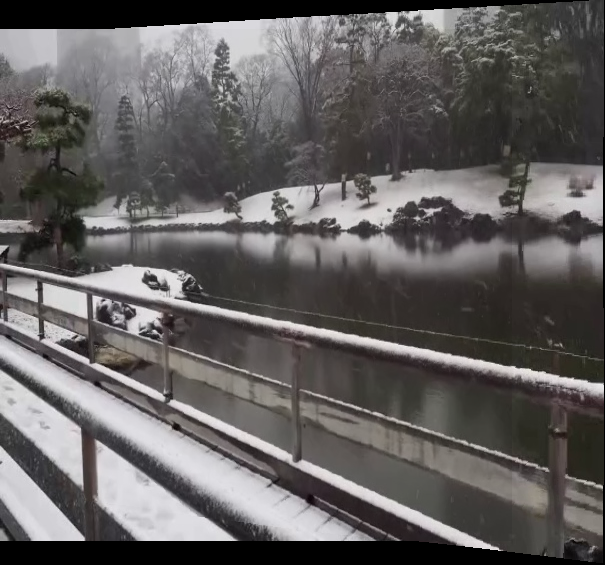}};
\spy [closeup_w,magnification=2.5] on ($(FigA)+(0.00, -0.0625)$) 
    in node[largewindow_w,anchor=east]       at ($(FigA.north) + (0.33,-0.61)$);
\spy [closeup_w,magnification=3.5] on ($(FigA)+(-0.21, -0.035)$) 
    in node[largewindow_w,anchor=east]       at ($(FigA.north) + (-0.17,-0.61)$);
\end{tikzpicture}
\begin{tikzpicture}[x=6cm, y=6cm, spy using outlines={every spy on node/.append style={smallwindow_w}}]
\node[anchor=south] (FigA) at (0,0) {\includegraphics[height=1.62in,width=1.62in]{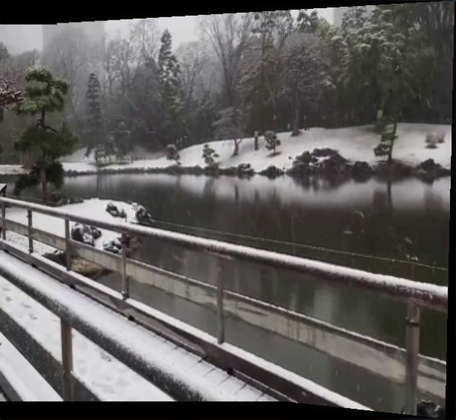}};
\spy [closeup_w,magnification=2.5] on ($(FigA)+(0.00, -0.065)$) 
    in node[largewindow_w,anchor=east]       at ($(FigA.north) + (0.33,-0.61)$);
\spy [closeup_w,magnification=3.5] on ($(FigA)+(-0.21, -0.04)$)  
    in node[largewindow_w,anchor=east]       at ($(FigA.north) + (-0.17,-0.61)$);
\end{tikzpicture}
\begin{tikzpicture}[x=6cm, y=6cm, spy using outlines={every spy on node/.append style={smallwindow_w}}]
\node[anchor=south] (FigA) at (0,0) {\includegraphics[height=1.62in,width=1.62in]{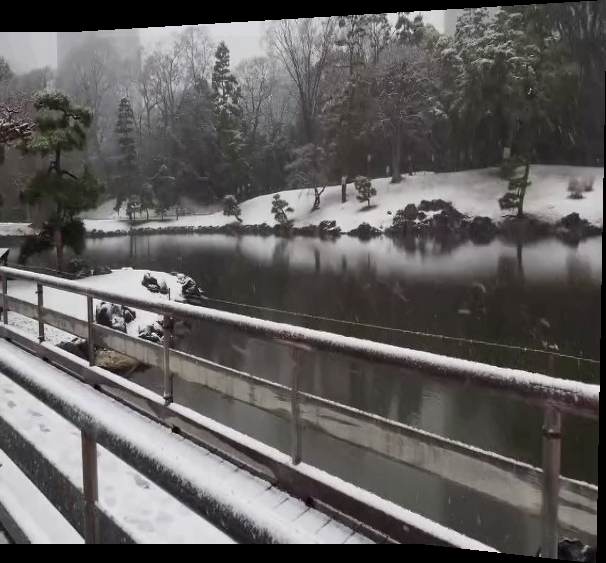}};
\spy [closeup_w,magnification=2.5] on ($(FigA)+(0.00, -0.065)$) 
    in node[largewindow_w,anchor=east]       at ($(FigA.north) + (0.33,-0.61)$);
\spy [closeup_w,magnification=3.5] on ($(FigA)+(-0.21, -0.04)$)  
    in node[largewindow_w,anchor=east]       at ($(FigA.north) + (-0.17,-0.61)$);
\end{tikzpicture}
\\
\vspace{-1mm}

\begin{tikzpicture}[x=6cm, y=6cm, spy using outlines={every spy on node/.append style={smallwindow_b}}]
\node[anchor=south] (FigA) at (0,0) {\includegraphics[height=1.62in,width=1.62in]{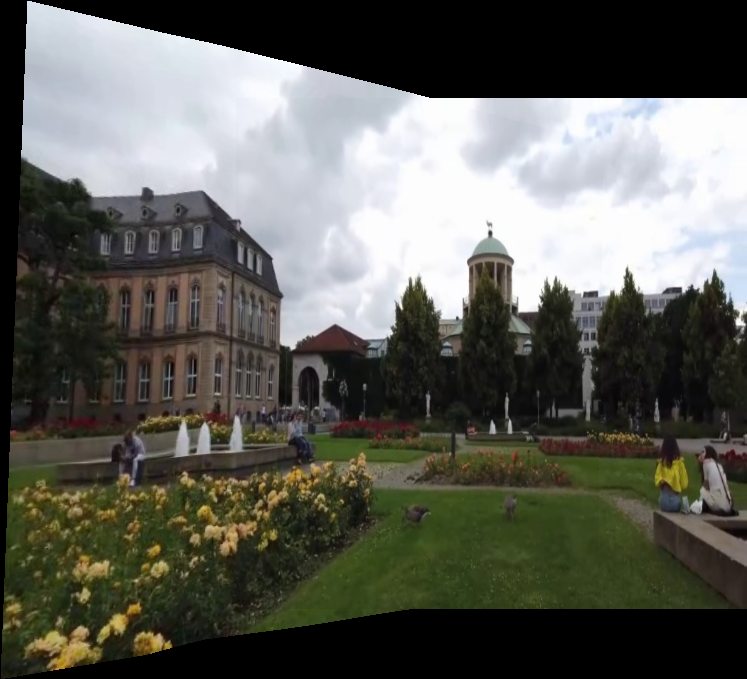}};
\spy [closeup_b,magnification=3] on ($(FigA)+(0.107, 0.068)$) 
    in node[largewindow_b,anchor=east]       at ($(FigA.north) + (0.33,-0.11)$);
\spy [closeup_b,magnification=4] on ($(FigA)+(0.13, -0.125)$)  
    in node[largewindow_b,anchor=east]       at ($(FigA.north) + (-0.17,-0.61)$);
\end{tikzpicture}
\begin{tikzpicture}[x=6cm, y=6cm, spy using outlines={every spy on node/.append style={smallwindow_b}}]
\node[anchor=south] (FigA) at (0,0) {\includegraphics[height=1.62in,width=1.62in]{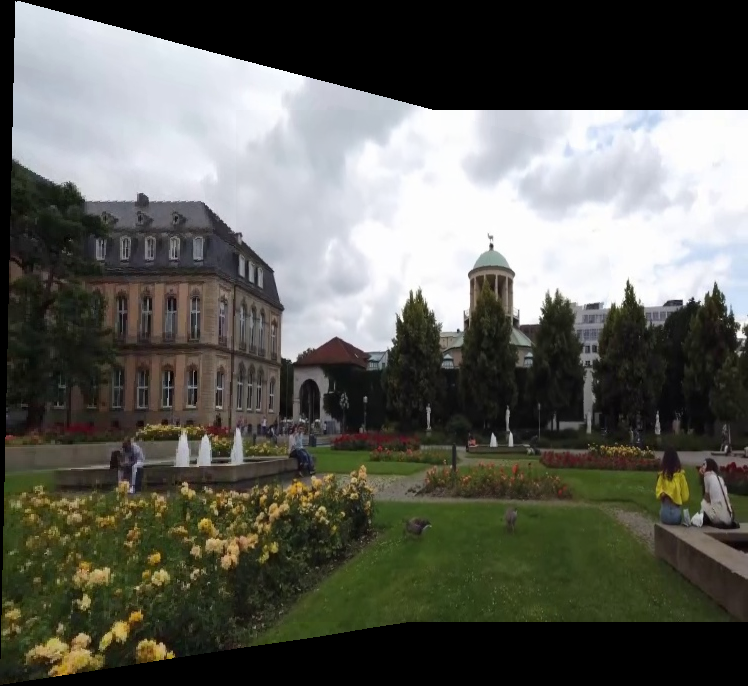}};
\spy [closeup_b,magnification=3] on ($(FigA)+(0.107, 0.06)$) 
    in node[largewindow_b,anchor=east]       at ($(FigA.north) + (0.33,-0.11)$);
\spy [closeup_b,magnification=4] on ($(FigA)+(0.13, -0.135)$)  
    in node[largewindow_b,anchor=east]       at ($(FigA.north) + (-0.17,-0.61)$);
\end{tikzpicture}
\begin{tikzpicture}[x=6cm, y=6cm, spy using outlines={every spy on node/.append style={smallwindow_b}}]
\node[anchor=south] (FigA) at (0,0) {\includegraphics[height=1.62in,width=1.62in]{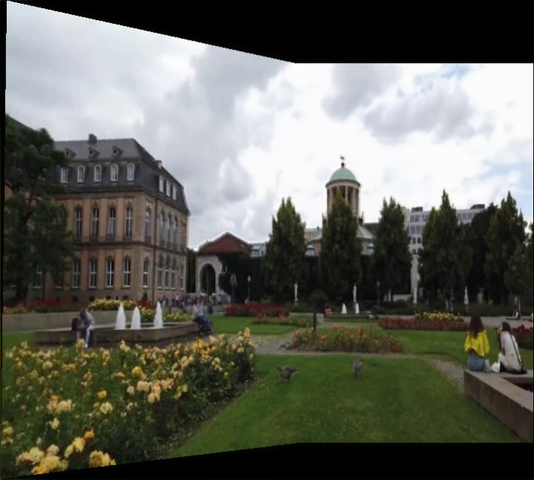}};
\spy [closeup_b,magnification=3] on ($(FigA)+(0.097, 0.07)$)  
    in node[largewindow_b,anchor=east]       at ($(FigA.north) + (0.33,-0.11)$);
\spy [closeup_b,magnification=3.5] on ($(FigA)+(0.122, -0.14)$)  
    in node[largewindow_b,anchor=east]       at ($(FigA.north) + (-0.17,-0.61)$);
\end{tikzpicture}
\begin{tikzpicture}[x=6cm, y=6cm, spy using outlines={every spy on node/.append style={smallwindow_b}}]
\node[anchor=south] (FigA) at (0,0) {\includegraphics[height=1.62in,width=1.62in]{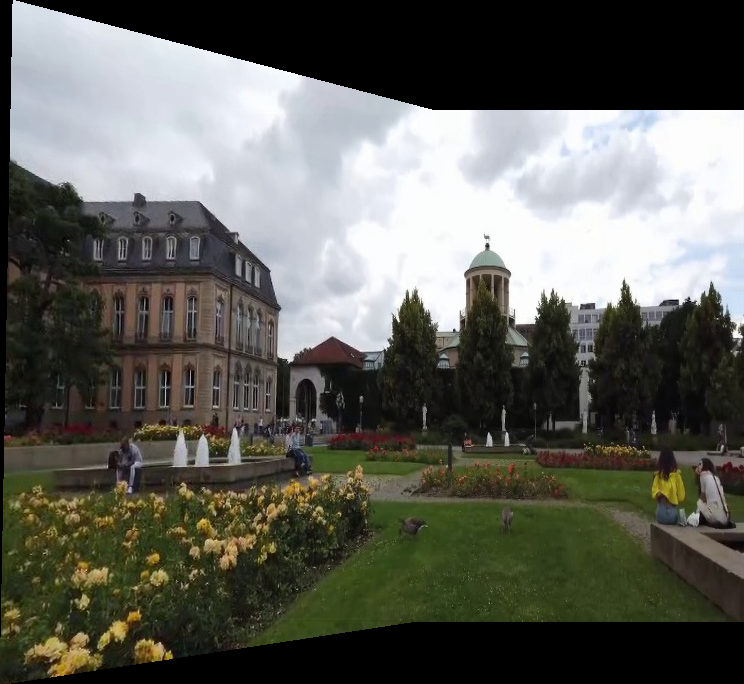}};
\spy [closeup_b,magnification=3] on ($(FigA)+(0.106, 0.06)$) 
    in node[largewindow_b,anchor=east]       at ($(FigA.north) + (0.33,-0.11)$);
\spy [closeup_b,magnification=4] on ($(FigA)+(0.128, -0.135)$)  
    in node[largewindow_b,anchor=east]       at ($(FigA.north) + (-0.17,-0.61)$);
\end{tikzpicture}
\\
\vspace{-1mm}

\begin{tikzpicture}[x=6cm, y=6cm, spy using outlines={every spy on node/.append style={smallwindow_w}}]
\node[anchor=south] (FigA) at (0,0) {\includegraphics[height=1.62in,width=1.62in]{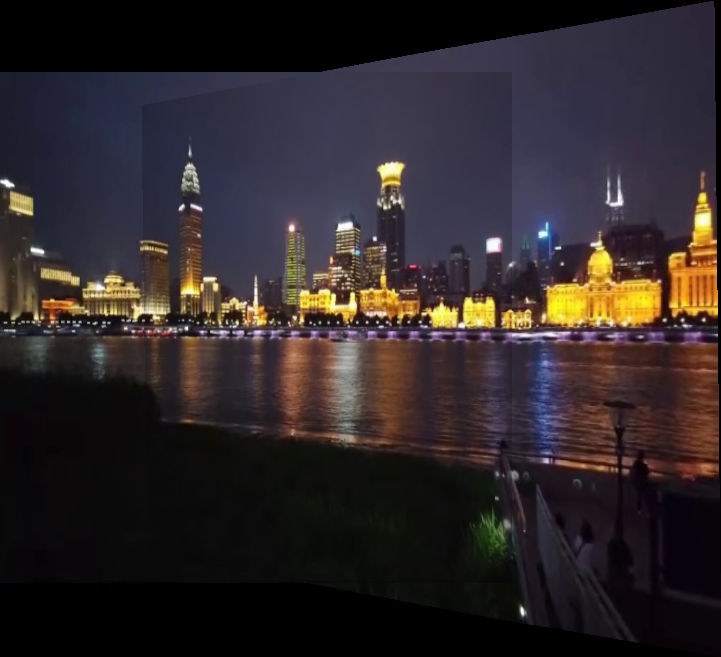}};
\spy [closeup_w,magnification=3] on ($(FigA)+(0.015, 0.027)$) 
    in node[largewindow_w,anchor=east]       at ($(FigA.north) + (0.33,-0.61)$);
\spy [closeup_w,magnification=3] on ($(FigA)+(-0.195,0.078)$)  
    in node[largewindow_w,anchor=east]       at ($(FigA.north) + (-0.17,-0.61)$);
\end{tikzpicture}
\begin{tikzpicture}[x=6cm, y=6cm, spy using outlines={every spy on node/.append style={smallwindow_w}}]
\node[anchor=south] (FigA) at (0,0) {\includegraphics[height=1.62in,width=1.62in]{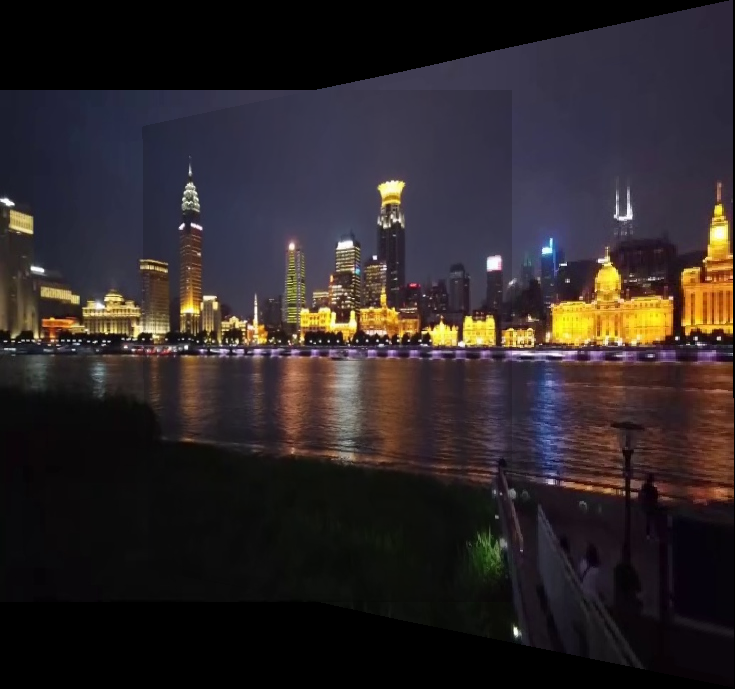}};
\spy [closeup_w,magnification=3] on ($(FigA)+(0.008, 0.024)$) 
    in node[largewindow_w,anchor=east]       at ($(FigA.north) + (0.33,-0.61)$);
\spy [closeup_w,magnification=3] on ($(FigA)+(-0.20,0.07)$)  
    in node[largewindow_w,anchor=east]       at ($(FigA.north) + (-0.17,-0.61)$);
\end{tikzpicture}
\begin{tikzpicture}[x=6cm, y=6cm, spy using outlines={every spy on node/.append style={smallwindow_w}}]
\node[anchor=south] (FigA) at (0,0) {\includegraphics[height=1.62in,width=1.62in]{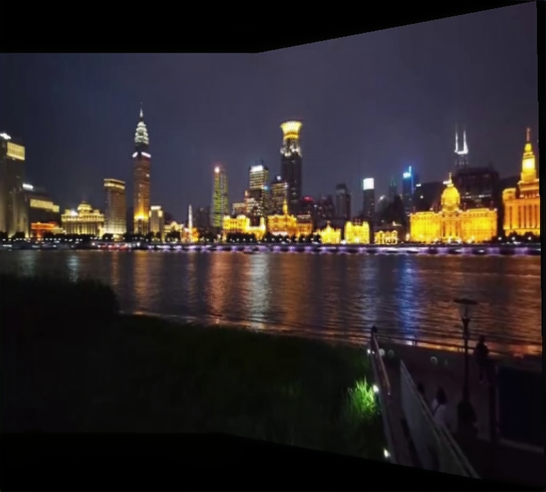}};
\spy [closeup_w,magnification=3.0] on ($(FigA)+( 0.012, 0.029)$) 
    in node[largewindow_w,anchor=east]       at ($(FigA.north) + (0.33,-0.61)$);
\spy [closeup_w,magnification=3] on ($(FigA)+(-0.20,0.08)$)  
    in node[largewindow_w,anchor=east]       at ($(FigA.north) + (-0.17,-0.61)$);
\end{tikzpicture}
\begin{tikzpicture}[x=6cm, y=6cm, spy using outlines={every spy on node/.append style={smallwindow_w}}]
\node[anchor=south] (FigA) at (0,0) {\includegraphics[height=1.62in,width=1.62in]{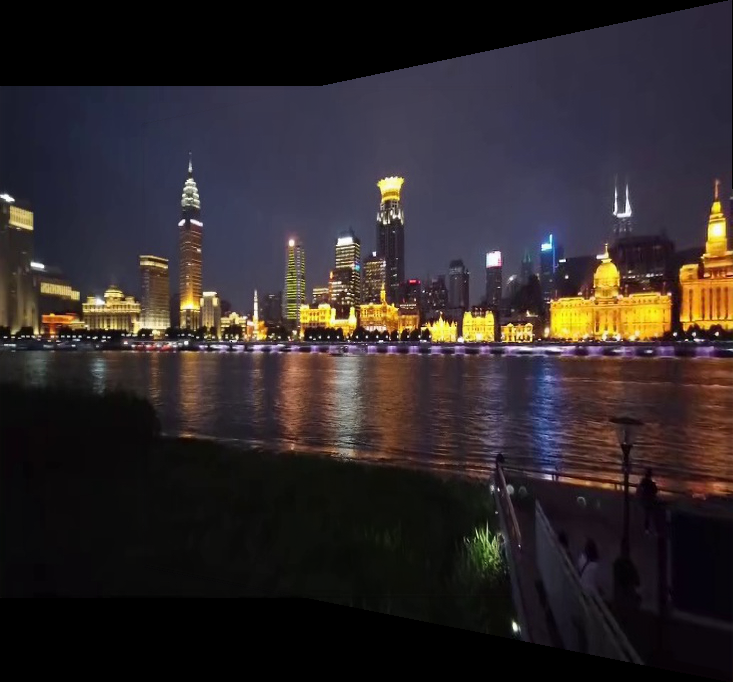}};
\spy [closeup_w,magnification=3.2] on ($(FigA)+(0.010, 0.0250)$) 
    in node[largewindow_w,anchor=east]       at ($(FigA.north) + (0.33,-0.61)$);
\spy [closeup_w,magnification=3] on ($(FigA)+(-0.20,0.072)$)  
    in node[largewindow_w,anchor=east]       at ($(FigA.north) + (-0.17,-0.61)$);
\end{tikzpicture}
\\
\vspace{-1mm}
\hspace{-1.2mm}
\stackunder[2pt]{
\begin{tikzpicture}[x=6cm, y=6cm, spy using outlines={every spy on node/.append style={smallwindow_w}}]
\node[anchor=south] (FigA) at (0,0) {\includegraphics[height=1.62in,width=1.62in]{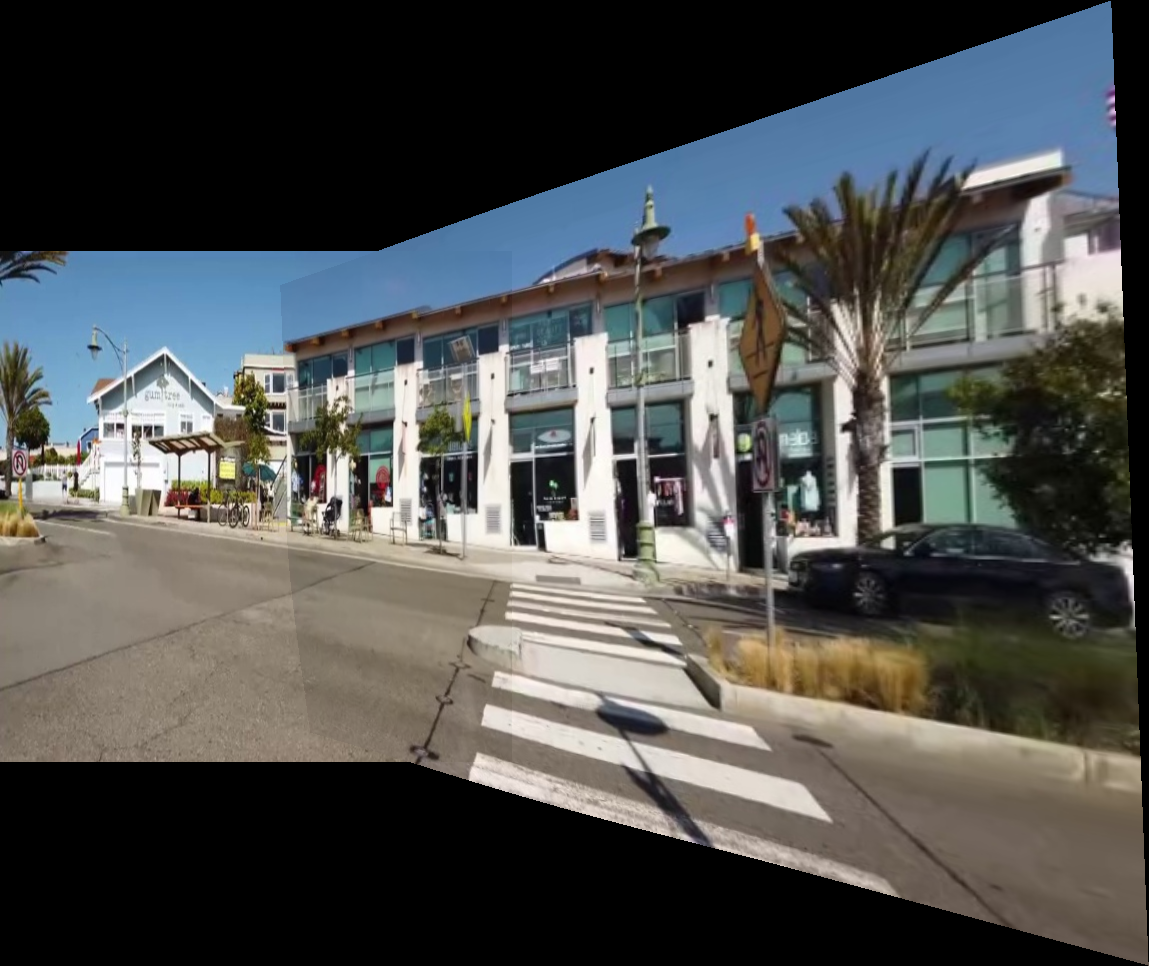}};
\spy [closeup_w,magnification=6] on ($(FigA)+(0.140, -0.005)$) 
    in node[largewindow_w,anchor=east]       at ($(FigA.north) + (0.33,-0.61)$);
\spy [closeup_w,magnification=6.5] on ($(FigA)+(-0.067,0.094)$)  
    in node[largewindow_w,anchor=east]       at ($(FigA.north) + (-0.17,-0.61)$);
\end{tikzpicture}}
{Robust ELA \cite{li2017parallax}}
\hspace{-1.32mm}
\stackunder[2pt]{
\begin{tikzpicture}[x=6cm, y=6cm, spy using outlines={every spy on node/.append style={smallwindow_w}}]
\node[anchor=south] (FigA) at (0,0) {\includegraphics[height=1.62in,width=1.62in]{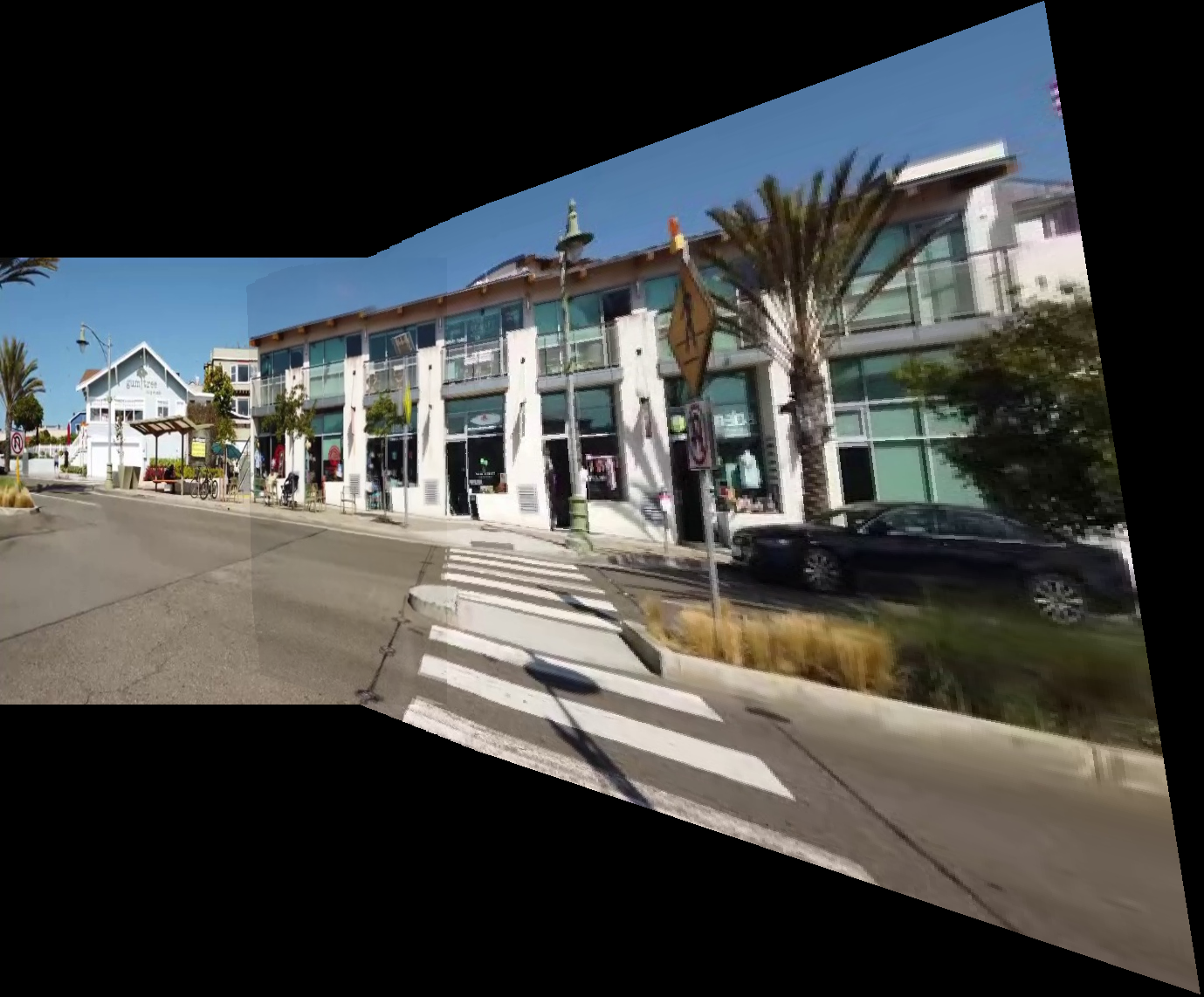}};
\spy [closeup_w,magnification=6] on ($(FigA)+(0.083, 0.021)$) 
    in node[largewindow_w,anchor=east]       at ($(FigA.north) + (0.33,-0.61)$);
\spy [closeup_w,magnification=7.5] on ($(FigA)+(-0.112,0.105)$)  
    in node[largewindow_w,anchor=east]       at ($(FigA.north) + (-0.17,-0.61)$);
\end{tikzpicture}}
{APAP \cite{zaragoza2013projective}}
\hspace{-1.31mm}
\stackunder[2pt]{
\begin{tikzpicture}[x=6cm, y=6cm, spy using outlines={every spy on node/.append style={smallwindow_w}}]
\node[anchor=south] (FigA) at (0,0) {\includegraphics[height=1.62in,width=1.62in]{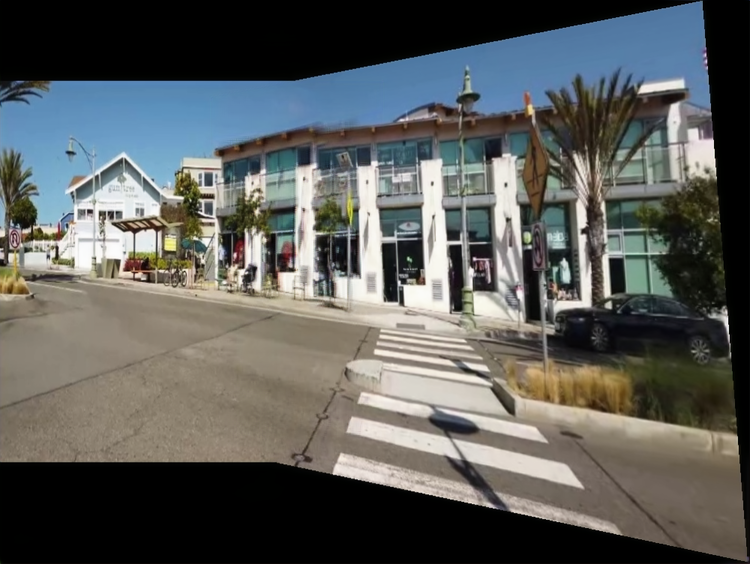}};
\spy [closeup_w,magnification=6] on ($(FigA)+(0.173, 0.014)$) 
    in node[largewindow_w,anchor=east]       at ($(FigA.north) + (0.33,-0.61)$);
\spy [closeup_w,magnification=5.5] on ($(FigA)+(-0.027,0.147)$)  
    in node[largewindow_w,anchor=east]       at ($(FigA.north) + (-0.17,-0.61)$);
\end{tikzpicture}}
{UDIS \cite{nie2021unsupervised}}
\hspace{-1.4mm}
\stackunder[2pt]{
\begin{tikzpicture}[x=6cm, y=6cm, spy using outlines={every spy on node/.append style={smallwindow_w}}]
\node[anchor=south] (FigA) at (0,0) {\includegraphics[height=1.62in,width=1.62in]{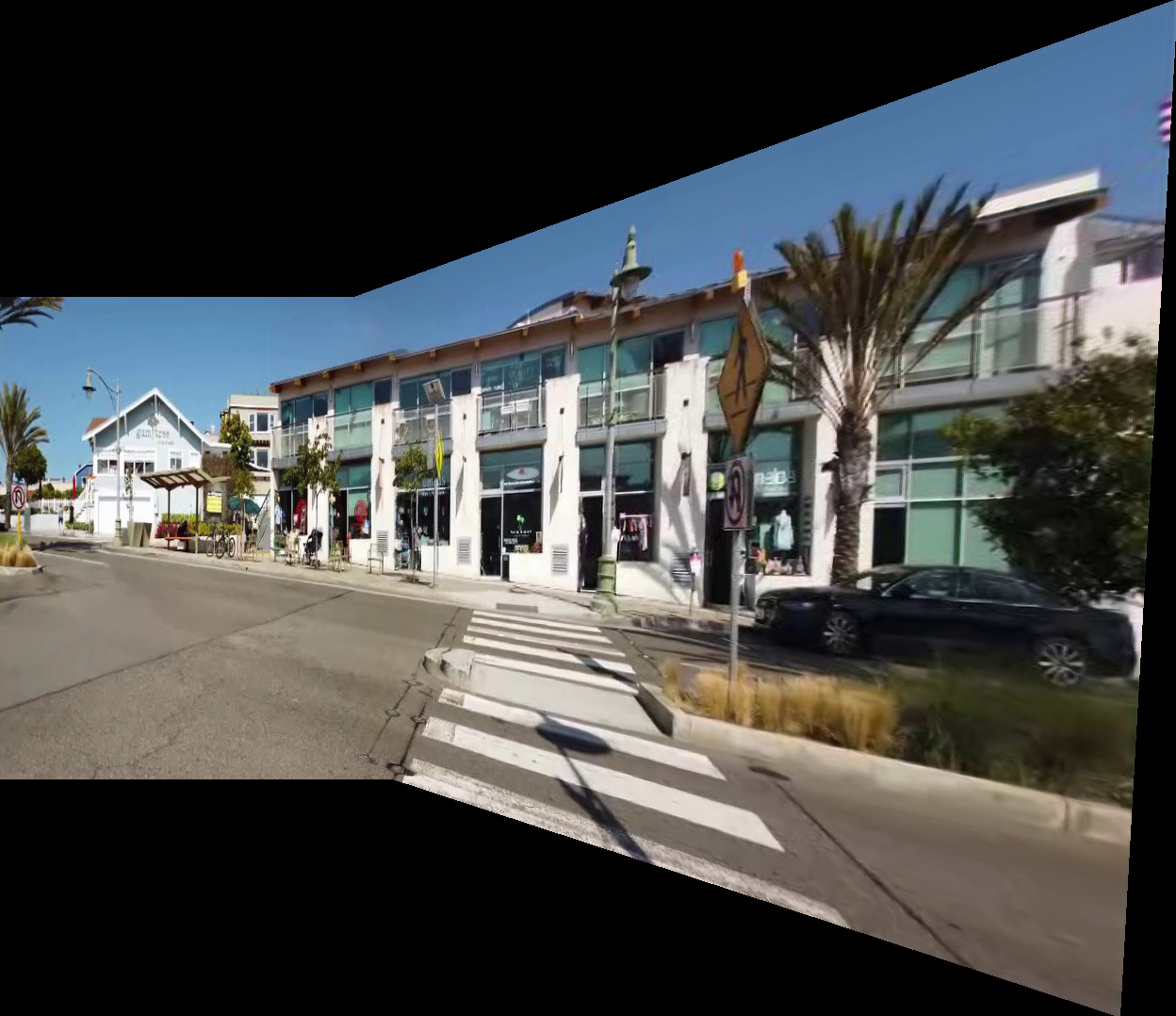}};
\spy [closeup_w,magnification=6.0] on ($(FigA)+(0.11, -0.014)$) 
    in node[largewindow_w,anchor=east]       at ($(FigA.north) + (0.33,-0.61)$);
\spy [closeup_w,magnification=7.5] on ($(FigA)+(-0.09,0.079)$)  
    in node[largewindow_w,anchor=east]       at ($(FigA.north) + (-0.17,-0.61)$);
\end{tikzpicture}}
{NIS (\textit{ours})}
\vspace*{-5pt}
\caption{Qualitative comparisons to other \textbf{image stitching approaches} on the UDIS-D real dataset \cite{nie2021unsupervised}.}
\vspace*{-10pt}
\label{fig:Qual_stit}
\end{figure*}

\section{Experiment}
\subsection{Dataset}
We use two datasets : MS-COCO \cite{lin2014microsoft} and UDIS-D \cite{nie2021unsupervised}. The first stage uses synthetic MS-COCO which is free from parallax errors. In the second stage, we use UDIS-D that includes various degrees of parallax errors.

\subsection{Implementation Details} \label{subsec: implementation}
\noindent{\bf Estimation of Alignment}
We employ a deep homography estimator IHN \cite{cao2022iterative} and robust ELA \cite{li2017parallax} for estimation of transformation to align images. We train a 2-scale IHN on UDIS-D and then use the estimated transformation to train NIS in our experiments. To check the performance of the trained model for unseen elastic warps, we use robust ELA to obtain aligning grids.

\noindent{\bf Neural Image Stitching}
The blender ($\boldsymbol{\it B_\eta}$) and the encoder ($\boldsymbol{\it E_\varphi}$) of the neural warping use EDSR \cite{lim2017enhanced} without upscaling module. The decoder ($\boldsymbol{\it G_\theta}$) is a 4-layer MLP with ReLUs, whose hidden dimension is 256. The amplitude, frequency, and phase estimators in neural warping are implemented with a single convolution layer without activation function. The amplitude and frequency estimators use a 256-channel $3\times3$ convolution layer and the phase estimation layer uses a 128-channel $1\times1$ convolution layer. 

\begin{figure}[t]
    \footnotesize
    \centering
    \begin{tikzpicture}
        \node(umb)[anchor=south west, inner sep=0pt]
            {\includegraphics[height=0.48\linewidth, width=0.48\linewidth]{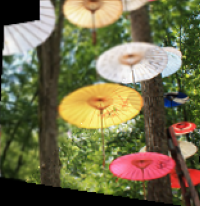}};
            \begin{scope}[x={(umb.south east)},y={(umb.north west)}]
              \foreach \i/\j in {{(0.003, 0.003) / (0.997, 0.997)}}
                \draw [white, thick] \i -- \j;
            \end{scope}
    \end{tikzpicture}
    \begin{tikzpicture}
        \node(zeb) [anchor=south west, inner sep=0pt]
            {\includegraphics[height=0.48\linewidth, width=0.48\linewidth]{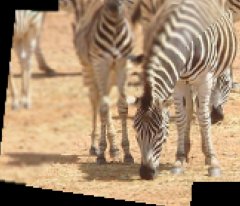}};
            \begin{scope}[x={(zeb.south east)},y={(zeb.north west)}]
              \foreach \i/\j in {{(0.003, 0.003) / (0.997, 0.997)}}
                \draw [white, thick] \i -- \j;
            \end{scope}
    \end{tikzpicture}
    \vspace{-5pt}
    \caption{\textbf{Qualitative comparison.} On the diagonal, the left side is the stitched image of UDIS \cite{nie2021unsupervised} and the right side is our image.}
    \label{fig:UDISvsOurs}
    \vspace{-15pt}
\end{figure}
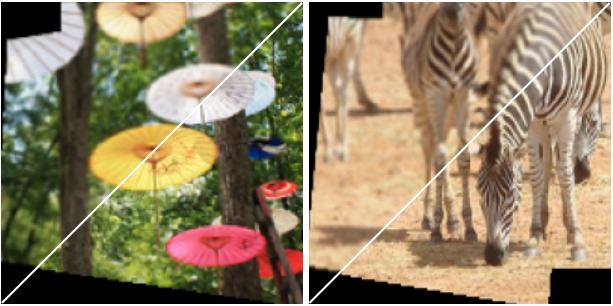

\subsection{Evaluation} \label{sec:eval}
\noindent\textbf{Quantitative result}
In \cref{tab:Quan_stit_syn}, we evaluate stitched image qualities on the synthetic MS-COCO dataset under synthesized ground-truth alignment. We find that the image stitching of UDIS negatively affects image quality as reported in \cref{tab:Quan_stit_syn}. Although the method employs bilinear warps, the UNet-based architectures cause such harmed imaging with jagging and blurred artifacts. However, the table validates that NIS successfully resolves the problem with 2.44, and 3.91 mPSNR gains compared to the bicubic warp, and bilinear warp, respectively. \cref{tab:Quan_stit_real} shows the summarized performances of image stitching quality. Since there is no ground truth for UDIS-D real dataset, we report NIQE \cite{mittal2012making}, PIQE \cite{venkatanath2015blind}, and BRISQUE \cite{mittal2012no}. We denote the fine-tuned NIS with (F). As shown in the table, the fine-tuned NIS that is capable of correcting color mismatches and parallax relaxing shows superior performances compared to the model without the training of feature blending. In addition, the comparison between UDIS and `LPC + Graph Cut' indicates that the learnable image stitching contributes the better image quality. Furthermore, the performance gains from a comparison of `IHN+NIS' to UDIS implies that our method for image stitching achieves significant image enhancement.

\noindent\textbf{Qualitative result}
Qualitative comparisons on UDIS-D real dataset are shown in \cref{fig:Qual_stit}. We compare NIS with the feature-based \cite{zaragoza2013projective}, \cite{li2017parallax} approaches, and a learning-based approach \cite{nie2021unsupervised}. APAP and robust ELA show parallax-tolerant image warping while UDIS and NIS blend parallax errors correcting the illumination difference. The qualitative comparisons of image qualities between UDIS and ours are provided in \cref{fig:UDISvsOurs}. The comparison supports the performance gains in \cref{tab:Quan_stit_syn} indicating that our method captures high-frequency details while keeping the image blending capability.

\begin{figure}[t]
    \centering
    \begin{subfigure}[b]{\columnwidth}
        \raisebox{0.17in}{\rotatebox{90}{Robust ELA}}
        \hspace{1mm}
        \includegraphics[width=2.55cm,height=2.55cm] {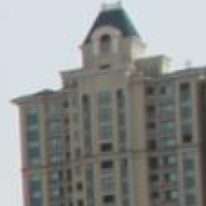}
        \includegraphics[width=2.55cm,height=2.55cm]{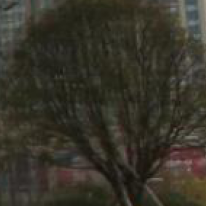}
        \includegraphics[width=2.55cm,height=2.55cm]{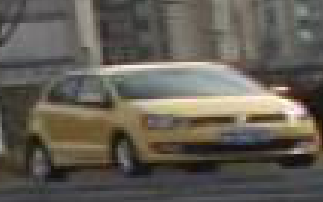}
        \\
        \raisebox{0.2in}{\rotatebox{90}{NIS (\textit{ours})}}
        \hspace{-0.6mm}
        \includegraphics[width=2.55cm,height=2.55cm]{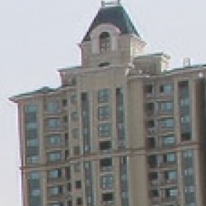}
        \includegraphics[width=2.55cm,height=2.55cm]{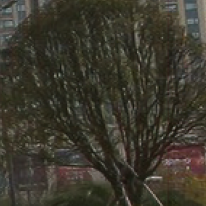}
        \includegraphics[width=2.55cm,height=2.55cm]{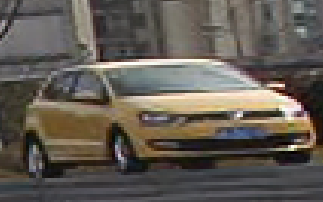}
    \end{subfigure}
    \vspace{-18pt}
    \caption{\textbf{Stitching with pre-computed Elastic Transformation grid.} Our pre-trained model is successfully applied on an elastically warped grid (unseen grid). We use robust ELA \cite{li2017parallax} to estimate the elastic grids.}
    \label{fig:ELA_and_ours}
\end{figure}

\noindent\textbf{Inference under Elastic warp}
Our model is trained under rigid transformations (or homography). To check if this training configuration can limit model performances for elastically transformed grids, we explore an experiment as in \cref{fig:ELA_and_ours}. As shown in the figure, we demonstrate an image aligned by a Thin-plate Spline grid and stitched by NIS. We use robust ELA \cite{li2017parallax} to estimate the elastic grid and generate the stitched image using NIS. As shown in the figure, our implicit neural representation for image stitching recovers high-frequency details on the totally unseen aligning grids.

\noindent\textbf{Cost Effectiveness}
In \cref{tab:time}, we compare the specifications of stitching methods for 3 different resolutions. To evaluate the model, we forward a common pair of images with 100 iterations, repeatedly. Then, we report max memory consumption (GB) and the average computation time (ms). As shown in the table, while NIS is cost-efficient for $192^2$ and $784^2$ sizes but, it 
shows weakness on high-resolution images $1536^2$ compared to the UDIS.


\subsection{Ablation Study}
\noindent \textbf{Fourier Features} To explore the contribution of Fourier features for image stitching on synthetic MS-COCO, we revisit the models with 3 configurations: 1) removal of amplitude, 2) frequency, and 3) phase estimator, respectively. We train all the cases with the first stage to clarify model performances on image enhancements. Note that the experiment using NIS without blending training shows different scale mPSNR as the model is fine-tuned on the other dataset, UDIS-D. In \cref{tab:Quan_stit_abl}, `w/o Amp.' removes the amplitude estimator by setting all the amplitudes of frequencies as 1. `w/o Freq.' estimates 128 frequencies whose size corresponds to half of the frequency estimator in NIS. The `w/o Phase' removes the phase estimator. As shown in \cref{tab:Quan_stit_abl}, the `w/o Amp.' model shows damaged Fourier features with significant performance degradation. The `w/o Freq.' model shows the reduced number of samples from the Fourier distribution. The result provided in the table emphasizes the importance of frequency priors. The `w/o Phase' model keeps the number of Fourier samples but shows a worse Fourier distribution. The observation has a negative effect as shown in the table.

\begin{table}[t]
    \small
    \begin{subtable}[t]{\linewidth}
        \centering
        \setlength{\tabcolsep}{1.5pt}
        \begin{tabular}{c
        |>{\centering\arraybackslash}p{0.25\textwidth}>{\centering\arraybackslash}p{0.25\textwidth}}
        \Xhline{2\arrayrulewidth}
            Method & mPSNR ($\uparrow$) & mSSIM ($\uparrow$) \\
            \hline
            w/o Amp. & 20.33 & 0.767 \\
            w/o Freq. & 43.23 & 0.990 \\
            w/o Phase & 44.07 & 0.990 \\
            NIS (\textit{ours} )& \textbf{44.16} & \textbf{0.991} \\
            \Xhline{2\arrayrulewidth}
        \end{tabular}
    \end{subtable}
    \vspace{-8pt}
    \caption{\textbf{Quantitative Ablation Study of NIS.}}
    \label{tab:Quan_stit_abl}
    \vspace{-5pt}
\end{table}

\begin{figure}[t]
    \scriptsize
    \centering
    \raisebox{0.22in}{\rotatebox{90}{\textit{Enhancement}}}
    \hspace{0.4mm}
    \stackunder[2pt]{\includegraphics[width=0.95in]{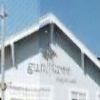}}{1000 iters}
    \stackunder[2pt]{\includegraphics[width=0.95in]{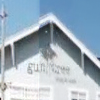}}{2000 iters}
    \stackunder[2pt]{\includegraphics[width=0.95in]{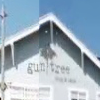}}{Best model} \\
    \vspace{2pt}
    \raisebox{0.19in}{\rotatebox{90}{\textit{Blending}}}
    \hspace{0.4mm}
    \stackunder[2pt]{\includegraphics[width=0.95in]{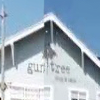}}{10 epoch}
    \stackunder[2pt]{\includegraphics[width=0.95in]{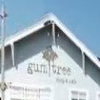}}{50 epoch}
    \stackunder[2pt]{\includegraphics[width=0.95in]{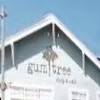}}{Best model}
    \vspace*{-8pt}
    \caption{\textbf{Ablation study of our training method.}}
    \label{fig:suboptimal}
    \vspace*{-10pt}
\end{figure}

\noindent \textbf{Learning Strategy} We investigate the contributions of our training methods in \cref{fig:suboptimal}. The first row contains learning procedures for enhanced feature reconstruction. The second row provides the observations during the fine-tuning stage. As in the figure, our learning methods are helpful for capturing high-frequency details and correcting color mismatches and misalignment. Despite the estimation of blended signals that may cause a worse image quality, we notice that our learning strategy for blended feature reconstruction provides visually pleasing stitched images.

\section{Discussion} \label{sec:dis}
\noindent {\bf Efficiency of Neural Warping} LTEW's pipeline for local INR prevents over/undershoot and blocky artifacts for more than $\times 8$ upscale factors. In contrast, our pipeline leverages displacement vector ($\mathbf{c}_m$) as a prior for the representation of stitched images with accelerated computations. In \cref{tab:cost}, we compare Neural Warping (NW) with LTEW, which is the state-of-the-art method for high-definition warp to verify our method. Following the recent research of arbitrary-scale super-resolution for \textit{Equirectangular projection} (ERP SR), we employ the same evaluation configurations as previous works \cite{deng2021lau, yoon2021spheresr}. The measurement of computation time is conducted under a fixed resolution to focus on complexity checking. After we train Neural Warping with $\mathbf{G}_\theta$ and LTE on ODI-SR $\times 4$ scale dataset, we test the models on ODI-SR test dataset \cite{deng2021lau} and SUN 360 \cite{xiao2012recognizing} dataset. As shown in \cref{tab:cost}, we see that our method shows favorable performances on both computational complexity and super-resolution.

\begin{figure}[t]
    \small
    \centering
    \stackunder[2pt]{\includegraphics[width=\linewidth]{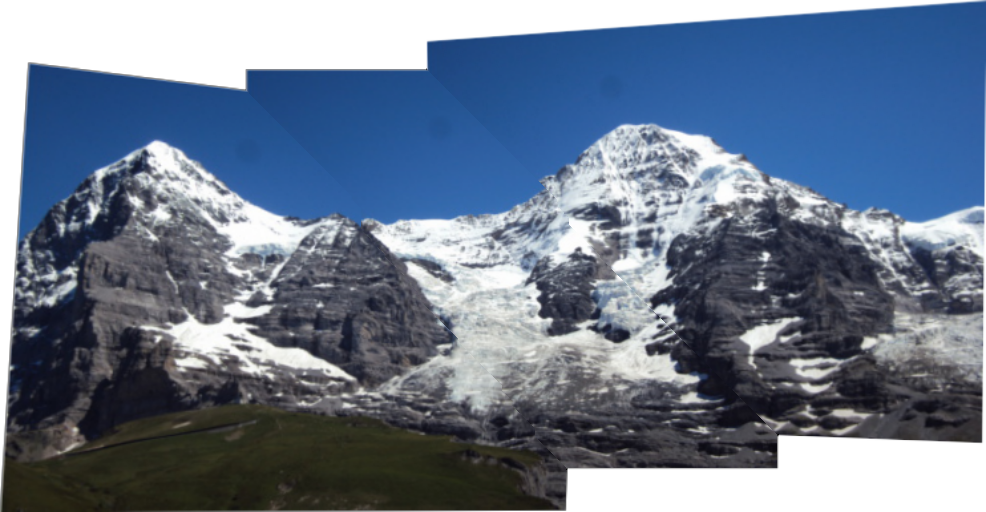}}{[Bilinear + Seam cutting] NIQE/PIQE/BRISQUE: 3.17/32.30/26.68}
    \vspace{5pt}

    \stackunder[2pt]{\includegraphics[width=\linewidth]{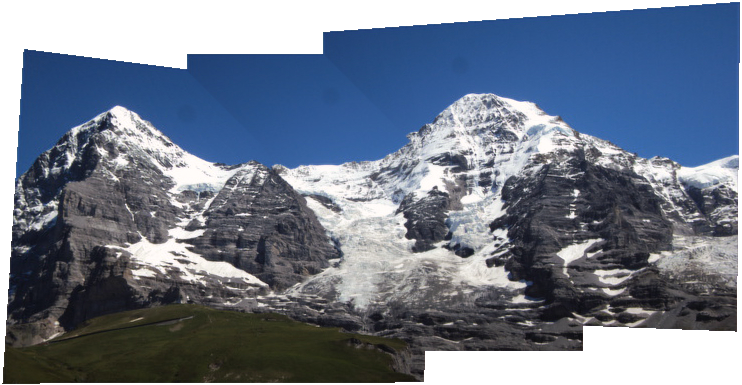}}{[NIS + Seam cutting] NIQE/PIQE/BRISQUE: \textcolor{red}{2.90}/\textcolor{red}{27.55}/\textcolor{red}{21.98}}
    \vspace{-4pt}
    \caption{\textbf{Comparison on the stitching of Multiple images.}}
    \label{fig:multiview}
    \vspace{-3pt}
\end{figure}

\noindent\textbf{NIS with Seam cutting} In \cref{fig:multiview}, we demonstrate a comparison of multi-view stitching to explore NIS with seam cutting. We prepare seam masks using \textit{OpenCV} `DpSeamFinder'. For image reconstruction with NIS, we determine a reference image $\mathbf{I}_r$ and two target images $\mathbf{I}_{t_1}, \mathbf{I}_{t_2}$. After that, using seam masks, we obtain two latent variables $\mathbf{C}'_1$ and $\mathbf{C}'_2$ from $\mathbf{z}_r$, $\mathbf{z}_{t_1}$ and $\mathbf{z}_r$, $\mathbf{z}_{t_2}$, respectively. Then NIS estimates the stitched image by blending $\mathbf{C}'_1, \mathbf{C}'_2$ and decoding the output. As shown in \cref{fig:multiview}, The enhancement of NIS demonstrates the potential to be used with seam cutting for advanced blending.

\section{Conclusion}
We proposed NIS, a novel end-to-end implicit neural reconstructor for image stitching. Our model predicted the dominant frequencies in warped domains from a pair of images to represent high-frequency details in a stitched image. Furthermore, our method successfully blended color mismatches and misalignments relaxing parallax errors. Our framework shows significant gains in synthetically stitched datasets over traditional methods for stitching, including bilinear and bicubic warps. Qualitative results on real images support that we successfully achieved high-frequency details for view-free image stitching compared to existing image stitching methods. On the other hand, we fail to simplify the training pipeline for neural image stitching. To design full end-to-end deep image stitching from image alignment to reconstruction, additional strategies for training would be required.

\begin{table}[t]
    \setlength{\tabcolsep}{1.2pt}
    \small
    \begin{subtable}[h]{\columnwidth}
        \centering
        \begin{tabular}{c
        |>{\centering\arraybackslash}p{2.5cm}
        |>{\centering\arraybackslash}p{2.0cm}
        >{\centering\arraybackslash}p{2.0cm}
        >{\centering\arraybackslash}p{2.0cm}}
        \hline
        \multirow{2}{*}{Dataset} & \multirow{2}{*}{Method} & \multicolumn{2}{c}{WS-PSNR} \\
         & & $\times8$ & $\times16$ \\
        \hline
        \multirow{2}{*}{ODI-SR} & LTEW & 25.53 & 23.91 \\
        & Neural Warping & \textbf{25.56} & \textbf{23.93} \\
        \hline
        \multirow{2}{*}{SUN 360} & LTEW & 25.60 & 23.59 \\
        & Neural Warping & \textbf{25.62} & \textbf{23.60} \\
        \hline
        \end{tabular}
        \subcaption{Performance on ERP SR.}
    \end{subtable}
    \vspace{5pt}

    \begin{subtable}[h]{\columnwidth}
        \centering
        \begin{tabular}{c
        |>{\centering\arraybackslash}p{2.1cm}|>{\centering\arraybackslash}p{2.1cm}|>{\centering\arraybackslash}p{2.3cm}>{\centering\arraybackslash}p{1.5cm}}
        \hline
        Size & Method & Memory (GB) & Time (ms) \\
        \hline
        \multirow{2}{*}{$256\times 256$} & LTEW & 0.36 & 20.50 \\
        & Neural Warping & 0.24 ($\mathbf{\textcolor{red}{\downarrow 33.3\%}}$) & 5.37 ($\mathbf{\textcolor{red}{\downarrow 73.8\%}}$) \\
        \hline
        \end{tabular}
        \subcaption{Cost-efficiency.}
    \end{subtable}
    \vspace*{-8pt}
    \caption{\textbf{Comparison of Neural Warping to LTEW.}}
    \label{tab:cost}
    \vspace{-15pt}
\end{table}


{\small
\bibliographystyle{ieee_fullname}
\bibliography{egbib}

\begin{thebibliography}{10}\itemsep=-1pt

\bibitem{agarwala2004interactive}
Aseem Agarwala, Mira Dontcheva, Maneesh Agrawala, Steven Drucker, Alex Colburn,
  Brian Curless, David Salesin, and Michael Cohen.
\newblock Interactive digital photomontage.
\newblock In {\em ACM SIGGRAPH 2004 Papers}, pages 294--302, 2004.

\bibitem{anderson2016jump}
Robert Anderson, David Gallup, Jonathan~T Barron, Janne Kontkanen, Noah
  Snavely, Carlos Hern{\'a}ndez, Sameer Agarwal, and Steven~M Seitz.
\newblock Jump: virtual reality video.
\newblock {\em ACM Transactions on Graphics (TOG)}, 35(6):1--13, 2016.

\bibitem{bay2006surf}
Herbert Bay, Tinne Tuytelaars, and Luc~Van Gool.
\newblock Surf: Speeded up robust features.
\newblock In {\em European conference on computer vision}, pages 404--417.
  Springer, 2006.

\bibitem{cao2022iterative}
Si-Yuan Cao, Jianxin Hu, Zehua Sheng, and Hui-Liang Shen.
\newblock Iterative deep homography estimation.
\newblock In {\em Proceedings of the IEEE/CVF Conference on Computer Vision and
  Pattern Recognition}, pages 1879--1888, 2022.

\bibitem{chen2021learning}
Yinbo Chen, Sifei Liu, and Xiaolong Wang.
\newblock Learning continuous image representation with local implicit image
  function.
\newblock In {\em Proceedings of the IEEE/CVF conference on computer vision and
  pattern recognition}, pages 8628--8638, 2021.

\bibitem{deng2021lau}
Xin Deng, Hao Wang, Mai Xu, Yichen Guo, Yuhang Song, and Li Yang.
\newblock Lau-net: Latitude adaptive upscaling network for omnidirectional
  image super-resolution.
\newblock In {\em Proceedings of the IEEE/CVF Conference on Computer Vision and
  Pattern Recognition}, pages 9189--9198, 2021.

\bibitem{detone2016deep}
Daniel DeTone, Tomasz Malisiewicz, and Andrew Rabinovich.
\newblock Deep image homography estimation.
\newblock {\em arXiv preprint arXiv:1606.03798}, 2016.

\bibitem{erlik2017homography}
Farzan Erlik~Nowruzi, Robert Laganiere, and Nathalie Japkowicz.
\newblock Homography estimation from image pairs with hierarchical
  convolutional networks.
\newblock In {\em Proceedings of the IEEE international conference on computer
  vision workshops}, pages 913--920, 2017.

\bibitem{fattal2002gradient}
Raanan Fattal, Dani Lischinski, and Michael Werman.
\newblock Gradient domain high dynamic range compression.
\newblock In {\em Proceedings of the 29th annual conference on Computer
  graphics and interactive techniques}, pages 249--256, 2002.

\bibitem{gao2011constructing}
Junhong Gao, Seon~Joo Kim, and Michael~S Brown.
\newblock Constructing image panoramas using dual-homography warping.
\newblock In {\em CVPR 2011}, pages 49--56. IEEE, 2011.

\bibitem{hartley2003multiple}
Richard Hartley and Andrew Zisserman.
\newblock {\em Multiple view geometry in computer vision}.
\newblock Cambridge university press, 2003.

\bibitem{hornik1989multilayer}
Kurt Hornik, Maxwell Stinchcombe, and Halbert White.
\newblock Multilayer feedforward networks are universal approximators.
\newblock {\em Neural networks}, 2(5):359--366, 1989.

\bibitem{jia2021leveraging}
Qi Jia, ZhengJun Li, Xin Fan, Haotian Zhao, Shiyu Teng, Xinchen Ye, and
  Longin~Jan Latecki.
\newblock Leveraging line-point consistence to preserve structures for wide
  parallax image stitching.
\newblock In {\em Proceedings of the IEEE/CVF conference on computer vision and
  pattern recognition}, pages 12186--12195, 2021.

\bibitem{jiang2022towards}
Zhiying Jiang, Zengxi Zhang, Xin Fan, and Risheng Liu.
\newblock Towards all weather and unobstructed multi-spectral image stitching:
  Algorithm and benchmark.
\newblock In {\em Proceedings of the 30th ACM International Conference on
  Multimedia}, pages 3783--3791, 2022.

\bibitem{kim2019deep}
Hak~Gu Kim, Heoun-Taek Lim, and Yong~Man Ro.
\newblock Deep virtual reality image quality assessment with human perception
  guider for omnidirectional image.
\newblock {\em IEEE Transactions on Circuits and Systems for Video Technology},
  30(4):917--928, 2019.

\bibitem{kingma2014adam}
Diederik~P Kingma and Jimmy Ba.
\newblock Adam: A method for stochastic optimization.
\newblock {\em arXiv preprint arXiv:1412.6980}, 2014.

\bibitem{lai2019video}
Wei-Sheng Lai, Orazio Gallo, Jinwei Gu, Deqing Sun, Ming-Hsuan Yang, and Jan
  Kautz.
\newblock Video stitching for linear camera arrays.
\newblock {\em arXiv preprint arXiv:1907.13622}, 2019.

\bibitem{le2020deep}
Hoang Le, Feng Liu, Shu Zhang, and Aseem Agarwala.
\newblock Deep homography estimation for dynamic scenes.
\newblock In {\em Proceedings of the IEEE/CVF Conference on Computer Vision and
  Pattern Recognition}, pages 7652--7661, 2020.

\bibitem{ltew-jaewon-lee}
Jaewon Lee, Kwang~Pyo Choi, and Kyong~Hwan Jin.
\newblock Learning local implicit fourier representation for image warping.
\newblock In {\em ECCV}, 2022.

\bibitem{lee2022local}
Jaewon Lee and Kyong~Hwan Jin.
\newblock Local texture estimator for implicit representation function.
\newblock In {\em Proceedings of the IEEE/CVF Conference on Computer Vision and
  Pattern Recognition}, pages 1929--1938, 2022.

\bibitem{li2017parallax}
Jing Li, Zhengming Wang, Shiming Lai, Yongping Zhai, and Maojun Zhang.
\newblock Parallax-tolerant image stitching based on robust elastic warping.
\newblock {\em IEEE Transactions on multimedia}, 20(7):1672--1687, 2017.

\bibitem{li2019attentive}
Jia Li, Yifan Zhao, Weihua Ye, Kaiwen Yu, and Shiming Ge.
\newblock Attentive deep stitching and quality assessment for $360^{\circ}$
  omnidirectional images.
\newblock {\em IEEE Journal of Selected Topics in Signal Processing},
  14(1):209--221, 2019.

\bibitem{liao2019single}
Tianli Liao and Nan Li.
\newblock Single-perspective warps in natural image stitching.
\newblock {\em IEEE transactions on image processing}, 29:724--735, 2019.

\bibitem{lim2017enhanced}
Bee Lim, Sanghyun Son, Heewon Kim, Seungjun Nah, and Kyoung Mu~Lee.
\newblock Enhanced deep residual networks for single image super-resolution.
\newblock In {\em Proceedings of the IEEE conference on computer vision and
  pattern recognition workshops}, pages 136--144, 2017.

\bibitem{lin2014microsoft}
Tsung-Yi Lin, Michael Maire, Serge Belongie, James Hays, Pietro Perona, Deva
  Ramanan, Piotr Doll{\'a}r, and C~Lawrence Zitnick.
\newblock Microsoft {COCO}: Common objects in context.
\newblock In {\em European conference on computer vision}, pages 740--755.
  Springer, 2014.

\bibitem{lin2011smoothly}
Wen-Yan Lin, Siying Liu, Yasuyuki Matsushita, Tian-Tsong Ng, and Loong-Fah
  Cheong.
\newblock Smoothly varying affine stitching.
\newblock In {\em CVPR 2011}, pages 345--352. IEEE, 2011.

\bibitem{lowe2004distinctive}
David~G Lowe.
\newblock Distinctive image features from scale-invariant keypoints.
\newblock {\em International journal of computer vision}, 60(2):91--110, 2004.

\bibitem{mildenhall2021nerf}
Ben Mildenhall, Pratul~P Srinivasan, Matthew Tancik, Jonathan~T Barron, Ravi
  Ramamoorthi, and Ren Ng.
\newblock Ne{RF}: Representing scenes as neural radiance fields for view
  synthesis.
\newblock {\em Communications of the ACM}, 65(1):99--106, 2021.

\bibitem{mittal2012no}
Anish Mittal, Anush~Krishna Moorthy, and Alan~Conrad Bovik.
\newblock No-reference image quality assessment in the spatial domain.
\newblock {\em IEEE Transactions on image processing}, 21(12):4695--4708, 2012.

\bibitem{mittal2012making}
Anish Mittal, Rajiv Soundararajan, and Alan~C Bovik.
\newblock Making a “completely blind” image quality analyzer.
\newblock {\em IEEE Signal processing letters}, 20(3):209--212, 2012.

\bibitem{nguyen2018unsupervised}
Ty Nguyen, Steven~W Chen, Shreyas~S Shivakumar, Camillo~Jose Taylor, and Vijay
  Kumar.
\newblock Unsupervised deep homography: A fast and robust homography estimation
  model.
\newblock {\em IEEE Robotics and Automation Letters}, 3(3):2346--2353, 2018.

\bibitem{nie2020view}
Lang Nie, Chunyu Lin, Kang Liao, Meiqin Liu, and Yao Zhao.
\newblock A view-free image stitching network based on global homography.
\newblock {\em Journal of Visual Communication and Image Representation},
  73:102950, 2020.

\bibitem{nie2021unsupervised}
Lang Nie, Chunyu Lin, Kang Liao, Shuaicheng Liu, and Yao Zhao.
\newblock Unsupervised deep image stitching: Reconstructing stitched features
  to images.
\newblock {\em IEEE Transactions on Image Processing}, 30:6184--6197, 2021.

\bibitem{nie2023learning}
Lang Nie, Chunyu Lin, Kang Liao, Shuaicheng Liu, and Yao Zhao.
\newblock Learning thin-plate spline motion and seamless composition for
  parallax-tolerant unsupervised deep image stitching.
\newblock {\em arXiv preprint arXiv:2302.08207}, 2023.

\bibitem{nie2020learning}
Lang Nie, Chunyu Lin, Kang Liao, and Yao Zhao.
\newblock Learning edge-preserved image stitching from large-baseline deep
  homography.
\newblock {\em arXiv preprint arXiv:2012.06194}, 2020.

\bibitem{perez2003poisson}
Patrick P{\'e}rez, Michel Gangnet, and Andrew Blake.
\newblock Poisson image editing.
\newblock In {\em ACM SIGGRAPH 2003 Papers}, pages 313--318, 2003.

\bibitem{porter1984compositing}
Thomas Porter and Tom Duff.
\newblock Compositing digital images.
\newblock In {\em Proceedings of the 11th annual conference on Computer
  graphics and interactive techniques}, pages 253--259, 1984.

\bibitem{rahaman2019spectral}
Nasim Rahaman, Aristide Baratin, Devansh Arpit, Felix Draxler, Min Lin, Fred
  Hamprecht, Yoshua Bengio, and Aaron Courville.
\newblock On the spectral bias of neural networks.
\newblock In {\em International Conference on Machine Learning}, pages
  5301--5310. PMLR, 2019.

\bibitem{rublee2011orb}
Ethan Rublee, Vincent Rabaud, Kurt Konolige, and Gary Bradski.
\newblock Orb: An efficient alternative to sift or surf.
\newblock In {\em 2011 International conference on computer vision}, pages
  2564--2571. Ieee, 2011.

\bibitem{son2021srwarp}
Sanghyun Son and Kyoung~Mu Lee.
\newblock {SRWarp}: Generalized image super-resolution under arbitrary
  transformation.
\newblock In {\em Proceedings of the IEEE/CVF conference on computer vision and
  pattern recognition}, pages 7782--7791, 2021.

\bibitem{song2022weakly}
Dae-Young Song, Geonsoo Lee, HeeKyung Lee, Gi-Mun Um, and Donghyeon Cho.
\newblock Weakly-supervised stitching network for real-world panoramic image
  generation.
\newblock In {\em European Conference on Computer Vision}, pages 54--71.
  Springer, 2022.

\bibitem{uyttendaele2001eliminating}
Matthew Uyttendaele, Ashley Eden, and Richard Skeliski.
\newblock Eliminating ghosting and exposure artifacts in image mosaics.
\newblock In {\em Proceedings of the 2001 IEEE Computer Society Conference on
  Computer Vision and Pattern Recognition. CVPR 2001}, volume~2, pages II--II.
  IEEE, 2001.

\bibitem{venkatanath2015blind}
N Venkatanath, D Praneeth, Maruthi~Chandrasekhar Bh, Sumohana~S Channappayya,
  and Swarup~S Medasani.
\newblock Blind image quality evaluation using perception based features.
\newblock In {\em 2015 Twenty First National Conference on Communications
  (NCC)}, pages 1--6. IEEE, 2015.

\bibitem{wang2020multi}
Lang Wang, Wen Yu, and Bao Li.
\newblock Multi-scenes image stitching based on autonomous driving.
\newblock In {\em 2020 IEEE 4th Information Technology, Networking, Electronic
  and Automation Control Conference (ITNEC)}, volume~1, pages 694--698. IEEE,
  2020.

\bibitem{wu2019gp}
Huikai Wu, Shuai Zheng, Junge Zhang, and Kaiqi Huang.
\newblock Gp-gan: Towards realistic high-resolution image blending.
\newblock In {\em Proceedings of the 27th ACM international conference on
  multimedia}, pages 2487--2495, 2019.

\bibitem{xiao2012recognizing}
Jianxiong Xiao, Krista~A Ehinger, Aude Oliva, and Antonio Torralba.
\newblock Recognizing scene viewpoint using panoramic place representation.
\newblock In {\em 2012 IEEE Conference on Computer Vision and Pattern
  Recognition}, pages 2695--2702. IEEE, 2012.

\bibitem{yoon2021spheresr}
Youngho Yoon, Inchul Chung, Lin Wang, and Kuk-Jin Yoon.
\newblock Sphere{SR}: 360deg image super-resolution with arbitrary projection
  via continuous spherical image representation.
\newblock In {\em Proceedings of the IEEE/CVF Conference on Computer Vision and
  Pattern Recognition}, pages 5677--5686, 2022.

\bibitem{zaragoza2013projective}
Julio Zaragoza, Tat-Jun Chin, Michael~S Brown, and David Suter.
\newblock As-projective-as-possible image stitching with moving {DLT}.
\newblock In {\em Proceedings of the IEEE conference on computer vision and
  pattern recognition}, pages 2339--2346, 2013.

\bibitem{zhang2014parallax}
Fan Zhang and Feng Liu.
\newblock Parallax-tolerant image stitching.
\newblock In {\em Proceedings of the IEEE Conference on Computer Vision and
  Pattern Recognition}, pages 3262--3269, 2014.

\bibitem{zhang2020content}
Jirong Zhang, Chuan Wang, Shuaicheng Liu, Lanpeng Jia, Nianjin Ye, Jue Wang, Ji
  Zhou, and Jian Sun.
\newblock Content-aware unsupervised deep homography estimation.
\newblock In {\em European Conference on Computer Vision}, pages 653--669.
  Springer, 2020.

\bibitem{zhang2020deep}
Lingzhi Zhang, Tarmily Wen, and Jianbo Shi.
\newblock Deep image blending.
\newblock In {\em Proceedings of the IEEE/CVF Winter Conference on Applications
  of Computer Vision}, pages 231--240, 2020.

\bibitem{zhao2021deep}
Yiming Zhao, Xinming Huang, and Ziming Zhang.
\newblock Deep {Lucas-Kanade} homography for multimodal image alignment.
\newblock In {\em Proceedings of the IEEE/CVF Conference on Computer Vision and
  Pattern Recognition}, pages 15950--15959, 2021.

\bibitem{zhou2019stn}
Qiang Zhou and Xin Li.
\newblock {STN-homography}: Direct estimation of homography parameters for
  image pairs.
\newblock {\em Applied Sciences}, 9(23):5187, 2019.

\end{thebibliography}
}

\end{document}